\documentclass[draft]{agujournal2019}
\usepackage{url} 
\usepackage{lineno}
\usepackage{soul}
\draftfalse
\journalname{Journal of Advances in Modeling Earth Systems (JAMES)}

\usepackage[pagebackref=false,hidelinks]{hyperref}
\hypersetup{
  colorlinks = false,    
  urlcolor   = blue,    
  linkcolor  = blue,    
  citecolor  = black     
}
\hypersetup{final}
\usepackage{natbib}

\usepackage{amsmath,amsfonts,bm}

\def\eqref#1{equation~\ref{#1}}

\def\1{\bm{1}}

\def\mA{{\bm{A}}}

\def\mM{{\bm{M}}}

\def\mY{{\bm{Y}}}

\DeclareMathAlphabet{\mathsfit}{\encodingdefault}{\sfdefault}{m}{sl}
\SetMathAlphabet{\mathsfit}{bold}{\encodingdefault}{\sfdefault}{bx}{n}
\newcommand{\tens}[1]{\bm{\mathsfit{#1}}}
\def\tA{{\tens{A}}}

\def\tX{{\tens{X}}}
\def\tY{{\tens{Y}}}

\def\sK{{\mathbb{K}}}

\usepackage{url}
\usepackage{amsmath}
\usepackage{booktabs} 

\usepackage{graphicx} 
\usepackage{amsfonts} 
\usepackage{amsmath,soul}
\usepackage{amssymb}
\usepackage[capitalize]{cleveref} 
\crefname{appsubsec}{Appendix}{Appendices}
\Crefname{appsubsec}{Appendix}{Appendices}
\crefname{appsec}{}{}
\Crefname{appsec}{}{}

\usepackage{dsfont} 
\usepackage[font=small,skip=0pt]{subcaption} 
\usepackage{balance}
\usepackage{bm} 
\usepackage{wrapfig}
\usepackage{float} 

\usepackage{makecell}

\newcommand{\cmmnt}[1]{\ignorespaces}

\newcommand{\bit}{\begin{itemize}}
\newcommand{\ei}{\end{itemize}}

\makeatletter
\renewcommand\paragraph{\@startsection{subsubsection}{4}{\z@}%
{0.25ex \@plus.5ex \@minus.2ex}%
{-.15em}%
{\normalfont\normalsize\itshape}}
\makeatother

\usepackage[english]{babel}
\usepackage[autostyle, english = american]{csquotes}
\MakeOuterQuote{"}

\newcommand{\olive}{\textcolor{olive}}

\usepackage[normalem]{ulem} 
\usepackage[normalem]{ulem} 
\newif\ifrevision
\newif\ifstrikingthrough
\usepackage{letltxmacro}
\LetLtxMacro\origcitep\citep
\renewcommand{\citep}[2][]{
  \ifstrikingthrough
    \mbox{\origcitep[#1]{#2}}%
  \else
    \origcitep[#1]{#2}%
  \fi
}
\LetLtxMacro\origcite\cite
\renewcommand{\cite}[2][]{%
  \ifstrikingthrough
    \mbox{\origcite[#1]{#2}}%
  \else
    \origcite[#1]{#2}%
  \fi
}
\LetLtxMacro\origcitet\citet
\renewcommand{\citet}[2][]{%
  \ifstrikingthrough
    \mbox{\origcitet[#1]{#2}}%
  \else
    \origcitet[#1]{#2}%
  \fi
}
\LetLtxMacro\origcref\cref
\renewcommand{\cref}[1]{%
  \ifstrikingthrough
    \mbox{\origcref{#1}}%
  \else
    \origcref{#1}%
  \fi
}
\newcommand{\strikeolive}[1]{%
  \ifrevision
    \strikingthroughtrue
    \olive{\sout{#1}}%
    \strikingthroughfalse
  \fi
}
\newcommand{\blue}[1]{%
 \ifrevision
    {\color{blue}#1}
 \else
    {#1}%
 \fi
}
\revisionfalse 

\begin{document}

\title{MeltwaterBench: Deep learning for \\ spatiotemporal downscaling of surface meltwater}

\authors{Björn Lütjens\affil{1}, 
Patrick Alexander\affil{2}, 
Raf Antwerpen\affil{2},\\
Til Widmann\affil{3}, 
Guido Cervone\affil{4}, 
Marco Tedesco\affil{2}} 
\affiliation{1}{Department of Earth, Atmospheric, and Planetary Sciences, Massachusetts Institute of Technology}
\affiliation{2}{Lamont Doherty Earth Observatory, Columbia University}
\affiliation{3}{Massachusetts Institute of Technology}
\affiliation{4}{Institute for Computational and Data Sciences, Pennsylvania State University}

\correspondingauthor{Björn Lütjens, $\dagger$ current affiliation at IBM Research, }{lutjens@mit.edu} %

\justifying 
\tolerance=1600 

\begin{keypoints}
\item We create 100m resolution daily maps of surface meltwater over the Helheim Glacier, Greenland from 2017 to 2023 using deep learning
\item \blue{Fusing regional climate model projections with satellite data improves accuracy from 83\% to 95\%, evaluated against synthetic aperture radar targets}
\item We publish an open-source benchmark for assessing deep learning methods on spatiotemporal gap-filling
\end{keypoints}


\begin{abstract}
The Greenland ice sheet is melting at an accelerated rate due to processes that are not fully understood and hard to measure. The distribution of surface meltwater can help us understand these processes and is observable through remote sensing, but current maps of meltwater face a trade-off: They are either high-resolution in time or space, but not both. We develop a deep learning model that creates gridded surface meltwater maps at daily 100m resolution by fusing data streams from remote sensing observations and physics-based models. In particular, we spatiotemporally downscale regional climate model outputs using synthetic aperture radar (SAR), passive microwave, and a digital elevation model over the Helheim Glacier in Eastern Greenland from 2017-2023. The downscaled maps resolve fine-scale topographic structure and extreme melt events. And, using SAR-derived meltwater as  \strikeolive{``ground truth''}\blue{target data}, we show that a deep learning method that fuses all data streams is over 10 percentage points more accurate over our study area than existing non-deep learning approaches that only rely on regional climate model outputs (83\% {vs.} 95\% Acc.) or passive microwave observations (72\% vs. 95\% Acc.). We evaluate standard deep learning methods (UNet and DeepLabv3+), and publish our spatiotemporally aligned dataset as a benchmark, \textit{MeltwaterBench}, for intercomparisons with more complex data-driven downscaling methods. The code and data are available at \href{https://github.com/blutjens/meltwaterbench}{github.com/blutjens/meltwaterbench}. 
\end{abstract}

\section*{Plain language summary}
Understanding why the Greenland ice sheet has been melting faster is challenging due to the difficulty of observing the underlying processes. An important observable indicator is surface meltwater, which is water that forms on top of or within the first meters of the ice sheet. The highest resolution information on surface meltwater can be derived from satellites with a synthetic aperture radar (SAR) instrument, but the resulting data is hard to use due to temporal gaps from the satellites' flight paths. During such temporal gaps an extreme meltwater event that can produce billions of tons of meltwater within a single day could have occurred. To simplify the use of surface meltwater data, we propose a deep learning method that creates regularly-spaced, daily, high-resolution maps of surface meltwater. The deep learning model does so by fusing the information from SAR with other satellite data and physics-based simulations that are available on a daily basis. We show that surface meltwater maps from our deep learning model are significantly more accurate than currently used maps. And, to encourage the development of more complex models we publish our data as a benchmark dataset.

\section{Introduction}\label{sec:introduction}

The Greenland ice sheet (GrIS) is melting at its fastest rate in 12,000 years~\citep{Briner20meltingfastestsince12000yrs} and has contributed to approximately a quarter \strikeolive{4cm}of the past-century sea level rise~\citep{IPCC_2021_WGI_Ch_9}.
Projections of how such contribution will change in the future vary greatly \strikeolive{(5-33cm by 2100)}which is partially due to regional climate model uncertainty in ice mass loss processes and feedbacks~\citep{aschwanden19greenlandsealevelrise}.
Remote sensing instruments can help quantify ice mass loss processes, for example, by observing surface meltwater and linking it to atmospheric drivers as well as ice mass loss~\citep{mattingly18grisatmosphericriver,mattingly23grisfoehnwinds}. But, current observations of surface meltwater from different satellite instruments face shortcomings with respect to either spatial resolution, temporal coverage, or physical observability. We propose a deep learning-based downscaling method that can fuse remote sensing and modeled data into a daily high-resolution (100m) surface meltwater product to fill these data gaps\strikeolive{ and, eventually reduce uncertainties in the future GrIS sea level rise contribution}.

The primary spaceborne instruments for observing surface meltwater -- optical, microwave, and radar -- can miss localized and rapid ice mass loss processes, such as coastal melt events~\citep{noel17gicmelting}.
Optical remote sensing can detect surface water, but sub-surface water to a lesser extent, and is frequently obscured by cloud cover in Eastern Greenland~\citep{miles2017,hakkinen14greenland}. Passive microwave (PMW) observations occur daily and can penetrate clouds, but have a spatial resolution of $3.25{-}25$km~\citep{ashcraft2006comparison,colosio2020surface}. Coarsely resolved daily information is also available from in-situ point observations~\citep{fausto21promice} or regional climate model (RCM) reanalyses, such as the 5km Modèle Atmosphérique Régional v3.14 (MAR;~\citeauthor{grailet24marv314},~\citeyear{grailet24marv314}). Yet, coarse data can miss topographic features that lead to enhanced melting~\citep{noel2016}, is error-prone in coastal areas due to single pixels covering both land and ocean, and can miss important hydrological features such as crevasses, meltwater rivers, or lakes~\citep{vandeberg2020,noel2016,mcmillan2016}. Very high-resolution information on surface meltwater can be retrieved from cloud-penetrating radar, e.g., Sentinel-1 Synthetic Aperture Radar (SAR) since 2017 at $10$m, but the revisit time of 2-12 days can miss the onset of extreme melt events that can produce billions of tons of meltwater within a single day, as detailed in~\cref{sec:data_characteristics}. The low-frequency revisit also hinders a better understanding of the effect of rapid rainfall events, atmospheric rivers, or strong foehn winds on local supraglacial hydrology~\citep{broeke2023,bailey25atmosphericrivers,mattingly23grisfoehnwinds}. In summary, the coarse spatial or temporal resolution of data sources represents a barrier to understanding melting processes. 

A few downscaling methods exist for merging datasets in space and time, but deep learning has not yet been evaluated as a method for downscaling surface meltwater in Greenland. Instead, lower order functional fits have been used to produce daily 1km~\citep{noel2016} or 100m~\citep{tedesco23maranddem} surface meltwater maps in Greenland by combining RCM output and reanalyses with a static digital elevation model (DEM). Despite being valuable for understanding meltwater processes, efforts like these mostly focus on local interpolation approaches, such as linear regression, random forest, or gradient boosting fits in local (${\sim}3{\times}3$px) neighborhoods. 
\blue{This approach enhances spatial resolution~\citep{noel2016}, but cannot correct spatially coherent biases that extend beyond these neighborhoods, such as MAR's overestimation of melt near the ice-sheet margin~\citep{fettweis2011melting}. Deep learning methods, such as convolutional neural networks (CNNs), have a larger receptive field for correcting such large-scale biases and have been proposed for downscaling surface meltwater~\citep{husman24meltwater,husman24meltwaterjames}.}
\strikeolive{This approach enhances spatial resolution, but does not correct for larger-scale spatial biases in the input data streams~\citep{noel2016,tedesco23maranddem}. Deep learning methods, such as convolutional neural networks (CNNs), can correct large-scale biases and have been proposed for downscaling surface meltwater~\citep{husman24meltwater,husman24meltwaterjames}.} 
Most similar to our work, \citet{husman24meltwater} created 12-hourly $500$m maps of surface meltwater fraction over Antarctica by fusing remote sensing data (SAR and PMW) with a UNet, which is a common CNN-based encoder-decoder architecture~\citep{ronneberger15unet}. In comparison, our study focuses on Greenland and incorporates RCM simulations.

Downscaling spatiotemporal data is common across Earth system modeling with deep learning being seen as a promising technique~\citep{mardani25corrdiff}. Benchmarks that define an accessible dataset, metrics, and strong baselines are crucial to understanding if deep learning is achieving meaningful progress~\citep{lutjens25internalvariability,rasp20weatherbench}.
Benchmarks related to downscaling focus on super-resolution of single-image~\citep{agustsson17div2k,karras18celebahq,kurinchi21wisosuper}, multi-image~\citep{wolters23zooming}, or video data~\citep{liu14vid4kbenchmark}. But, most datasets in super-resolution use input and target imagery of the same modality and \blue{assume that biases are confined to a local neighborhood} \strikeolive{do not contain large-scale biases}, i.e., the regional average of high-resolution target pixels equals the value of the corresponding low-resolution input pixel~\citep{harder23physdownscaling}. In comparison, downscaling requires the correction of large-scale biases and translation across modalities, such as mapping PMW and DEM information onto surface meltwater. To the best of our knowledge, only two benchmarks exist for evaluating deep learning methods for downscaling:
\Citet{chen22rainnet} evaluates methods on downscaling global to regional precipitation reanalysis over the Eastern US, which does not capture the issue of translation across modalities. \Citet{langguth24downscalebench} released a dataset for downscaling atmospheric variables over the Alps, but it is missing baseline methods.

\begin{figure}[t]
  \centering
      \includegraphics[trim=0 0 0 0, width=0.98\columnwidth, angle = 0]{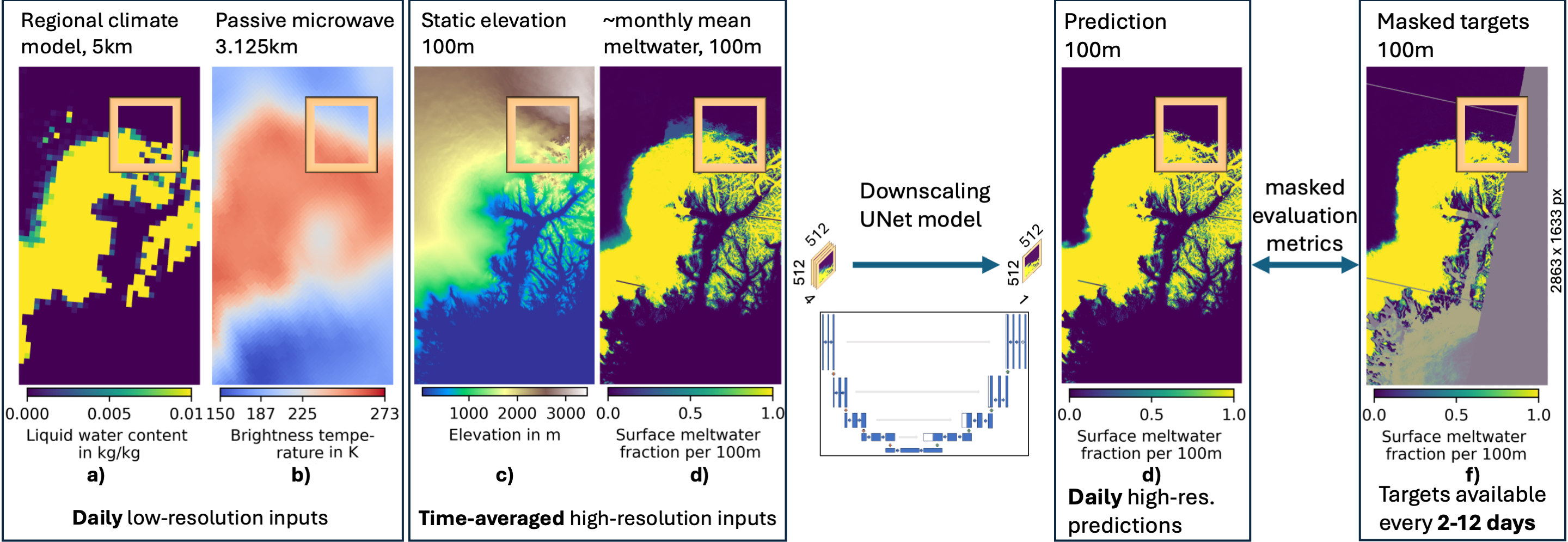}
\caption[]{\textbf{Overview.} Our benchmark, MeltwaterBench, poses a downscaling task, which is to predict a high-resolution Sentinel-1 SAR-derived map of surface meltwater fraction (f). These meltwater observations are only available every 2-12 days and are partially masked (f; gray area), due to the Sentinel-1 retrieval paths. The benchmark evaluates which machine learning method can most accurately predict meltwater maps, given low-resolution data from a regional climate model (a; MAR) and passive microwave observations (b; PMW). The available inputs also include a digital elevation model (c; DEM), a running mean of meltwater observations (d; time interpolate SAR), and optional auxiliary atmospheric variables and optical satellite observations (not shown; see~\cref{tab:mar_data_o}), which are not used in our baseline UNet. The images display June 24, 2018.}
\label{fig:overview}
\end{figure}

We propose a benchmark dataset, metrics, and baselines for evaluating deep learning methods on spatiotemporal downscaling of RCM-simulated surface meltwater over a study area surrounding the Helheim Glacier in Eastern Greenland, \blue{as outlined in~\cref{fig:overview}}.
For this study area, we derived 100m surface meltwater fraction for all available Sen-1 SAR observations using a well-validated threshold-based approach~\citep{ashcraft2006comparison}. With the surface meltwater targets, we created a machine learning (ML)-ready dataset from 2017-2023 with spatiotemporally aligned inputs from the 5km MAR regional climate model reanalysis, PMW, and DEM data, also shown in~\cref{fig:overview}.
We implement three UNet-based models as deep learning baselines and compare them to methods that rely on a single data stream and are representative of current practice.
As the best UNet achieves 95\% accuracy, we also create a daily high-resolution record of surface meltwater fraction over the Helheim Glacier for 2017-2023 .
\blue{Ultimately, such dense high-resolution meltwater products could help evaluate and constrain ice mass loss processes in regional climate models and reduce the spread in projections of runoff and GrIS sea level rise contribution.}

\section{Data and methods}\label{sec:data_and_methods}

\begin{figure}[ht]
  \centering
      \includegraphics[trim=0 0 0 0, width=0.78\columnwidth, angle = 0]{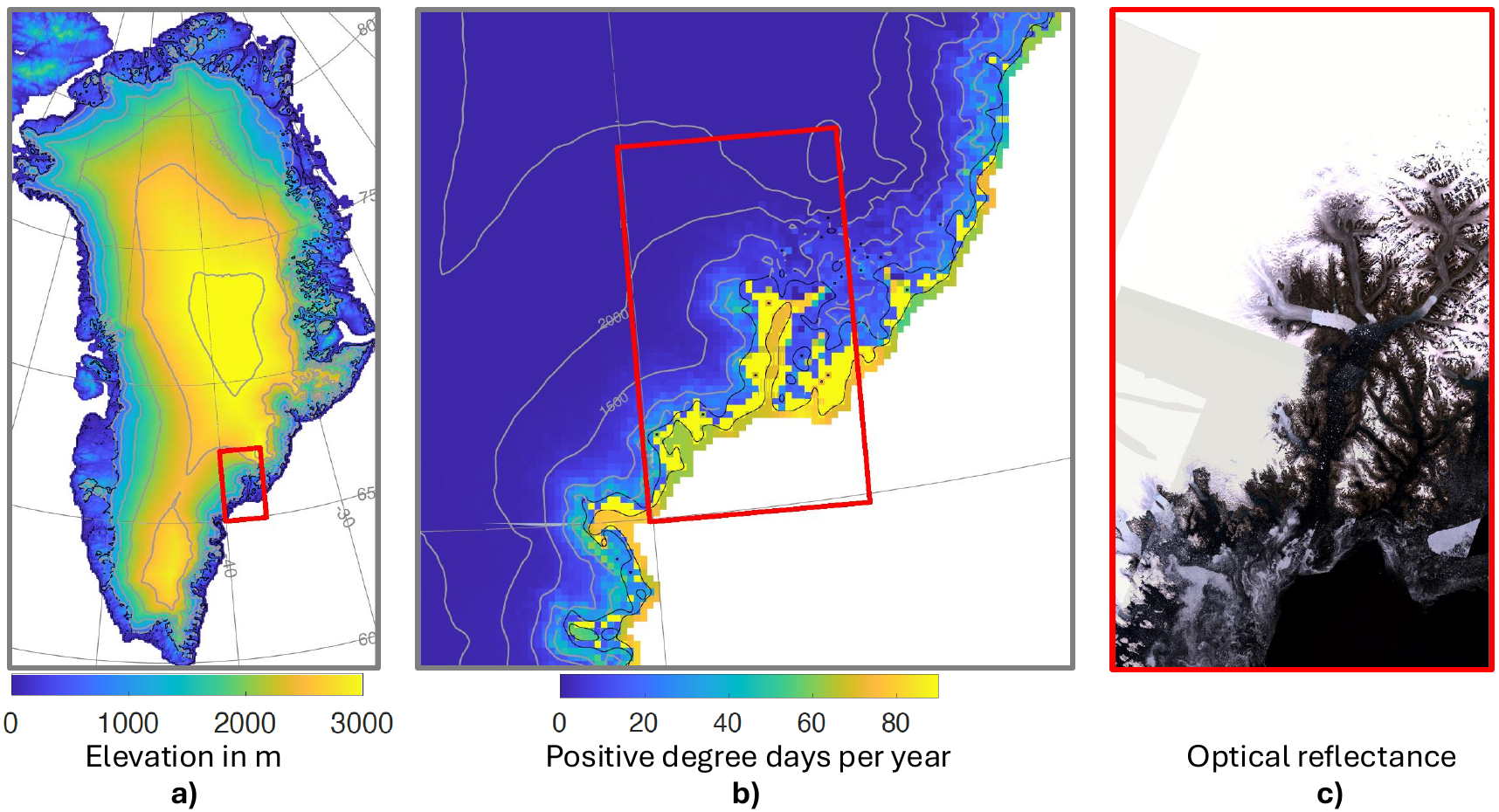}
\caption[]{\textbf{Study area.}~\Cref{fig:studyarea}a) shows the ice sheet surface elevation over Greenland and our study area (red rectangle) surrounding the Helheim glacier.~\Cref{fig:studyarea}b) shows the number of days per melting season (06/01-08/31) with near-surface air temperature ${>}0^\circ$C averaged over 2017-2023. \Cref{fig:studyarea}c) shows a satellite imagery mosaic from~\citep{howat17landsatmosaic}.}
\label{fig:studyarea}
\end{figure}

\subsection{Study area and period}\label{sec:study_area}
We focus on a region in Eastern Greenland centered around the Sermilik Fjord, Helheim Glacier (66.4°N, 38.2°W), as visualized in~\cref{fig:studyarea}. 
This marine-terminating glacier has been one of the largest contributors to the GrIS ice discharge~\citep{williams2021helheim}, 
and may be accelerated by an increase in surface meltwater~\citep{andresen2012rapid,Stevens_Nettles_Davis_Creyts_Kingslake_Ahlstrom_Larsen_2022}.
The region is suitable to evaluate downscaling techniques, due to the complex topography, variation between ice- and snow-covered sections, and availability of in-situ station data~\citep{shimada2016inter,antwerpen2022assessing, fausto21promice}.
Within the study area, we focus on the land because mass loss from grounded land ice is a major contributor to sea level rise~\citep{edwards2021projected}, and sea ice dynamics complicate SAR analysis~\citep{howell2019estimating}. We focus our analysis on 2017-2023, delineated by the start of Sentinel-1 SAR data collection and the end of the utilized RCM reanalysis.

\subsection{Data sources}\label{sec:data}

We download, reproject, and crop all data sources to a 100m Albers equal area projection over our study area, and save them as daily GeoTIFFs (reprojection details in~\cref{app:data}). We create a \textit{core} subset of the data that is used for all analyses, and publish a larger \textit{auxiliary} dataset to facilitate follow-up studies that may require a different selection of variables or time periods~\citep{HelheimData}.

\subsubsection{Synthetic aperture radar (SAR) data}
\begin{figure}[t]
  \centering
      \includegraphics[trim=0 0 0 0, width=0.98\columnwidth, angle = 0]{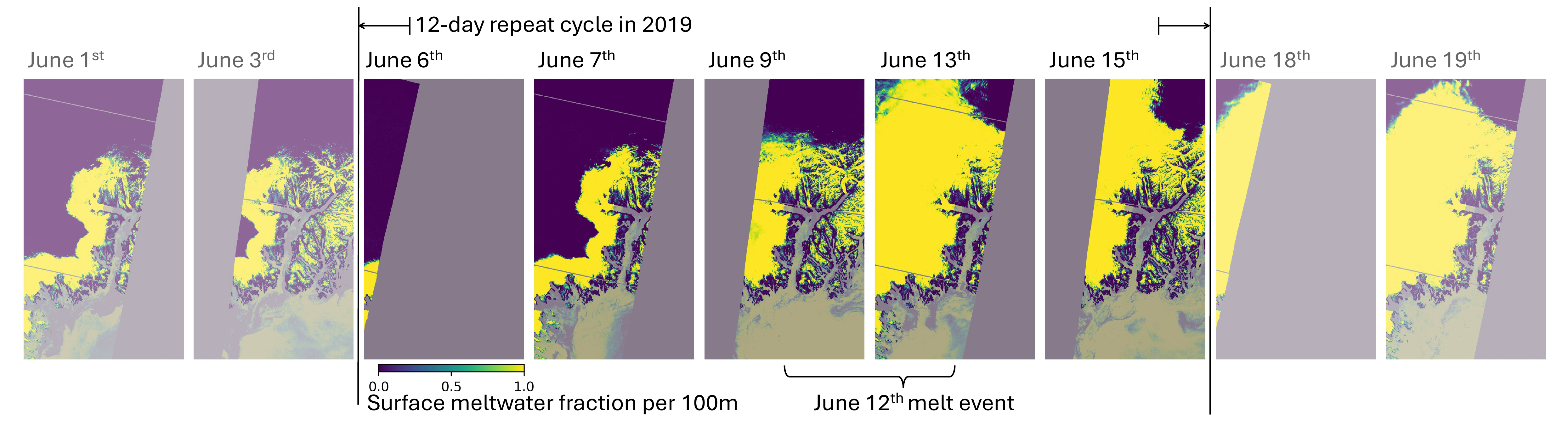}
\caption[]{\textbf{Surface meltwater targets.} The plots show Sentinel-1 synthetic aperture radar (SAR)-derived surface meltwater fraction per 100m grid cell during an extreme melt event on June 12, 2019 -- yellow indicates melt. During this event the melt area across the Greenland ice sheet almost doubled within a single day~\citep{tedesco2020unprecedented} and expanded rapidly across our northwestern study area. 
We mask out areas that are outside of satellite retrieval paths (gray blocks), over the ocean (southern study area), or impacted by post-processing artifacts (gray lines on June 13).} 
\label{fig:satellite_retrievel_paths_2019}
\end{figure}

~\Cref{fig:satellite_retrievel_paths_2019} illustrates the estimates of surface meltwater fraction per 100m grid cell that we derive from SAR, during an extreme melt event in 2019. To derive this surface meltwater fraction, we utilized SAR backscatter data from the European Space Agency Sentinel-1A and -1B (S1A and -B) satellites, distributed by NASA's Earth Observing System Data and Information System~\citep{Torres2017Sentinel1}. We used level-1 ground range detected data transmitted and collected in the horizontal polarization \blue{(HH)}, in interferometric wide swath (IW) mode in the C-band (5.405GHz)~\citep{torres12sen1}. The IW data has a spatial resolution of 5m by 20m and a swath width of 250km. Both satellites have a repeat cycle of 12 days. 
With multiple satellite tracks intersecting our study area and S1A and S1B being available during 2016-present and 2017-2021, respectively, SAR data partially covers our study area every 1-12 days. The retrieval path boundaries are visible as straight lines between gray and non-grayed area, e.g., in~\cref{fig:satellite_retrievel_paths_2019}. All S1A and -B data used in our final product were collected between 06:10 and 06:40 local solar time. 

Our SAR data workflow is depicted in~\cref{fig:sar_processing} and starts with a standard set of processing steps: orbital correction, subsetting, border noise removal, radiometric calibration, speckle filtering, and terrain correction. Then, we reproject the data onto a 10m resolution subset of the 100m Albers equal area grid. We convert backscattering intensity (in dB) to 10m binary melt by applying a threshold of -3dB relative to each year's winter mean backscatter. 
This \strikeolive{or a similar (-2.8dB)} threshold is commonly utilized~\citep{ashcraft2006comparison}\blue{, is} consistent with theory and observations \origcitep[e.g.][]{johnson2020evaluation,luckman2014surface,scher2021mapping,wismann2000monitoring,stiles1980active}\blue{, and has been specifically applied to Sentinel-1 SAR data to estimate glacier and ice sheet melt{~\citep{scher2021mapping,johnson2020evaluation,li2024glacier}}.} We mosaic all images from the same day into one map and, lastly, aggregate binary 10m melt onto the 100m grid by computing
the fraction of 10m cells exhibiting melt within each 100m grid cell.

\subsubsection{Passive microwave (PMW) data}
Satellite-based passive microwave (PMW) measurements are available from 1979 to the present and are commonly used to detect melt on the surface of ice sheets and glaciers~\origcitep[e.g.,][]{abdalati1995passive,liu2005wavelet,tedesco2007snowmelt,tedesco2009assessment,fettweis2011melting,colosio2020surface}. 
We sourced PMW brightness temperature observations from the Special Sensor Microwave Imager/Sounder (SSMIS) at 3.125km resolution using the MEaSUREs product with an effective resolution of 3.125-25km~\blue{\citep{Brodzik_data,rs12162552}}. The SSMIS observations are available every 12 hours -- mostly unaffected by local weather -- and we select each day's evening pass ($\approx$ 18:30 local solar time). We detail PMW in~\cref{app:pmw}. 

\subsubsection{Regional climate model data (MAR)}
We post-process 5km, daily-averaged data from the MARv3.14 regional climate model, which simulates atmospheric dynamics and resolves key processes regarding the ice sheet mass balance~\citep{grailet24marv314}. The MARv3.14 model incorporates observational data through being forced with six-hourly ERA5 reanalysis data~\citep{hersbach2020era5} and having been evaluated against remote sensing and in-situ datasets~\origcite[e.g.,][]{fettweis2011melting,fettweis2020grsmbmip,fettweis2021brief,delhasse2020brief}. We utilize the estimate of liquid water content within the first meter (WA1), which can be used analogous to surface meltwater presence~\citep{kittel2022assimilation,dethinne2023sensitivity}. 
We process and include additional MAR atmospheric variables related to surface meltwater, such as wind speed or shortwave downwelling solar radiation, in the auxiliary dataset. We list these variables in~\cref{tab:mar_data_o} and detail MARv3.14 in~\cref{app:mar}.

\subsubsection{Digital elevation model (DEM), land-ocean mask, and MODIS}\label{sec:dem}\label{sec:grimp}\label{sec:modis}

Besides dynamic variables, we add a static DEM mosaic at 100m that is generated from panchromatic stereoscopic imagery in 2008-2020 and lidar observations in summer 2019-2020~\blue{\citep{grimp2022data,howat2014greenland}}, and is detailed in~\cref{app:dem}. We also use a static 100m land-ocean mask, which is derived from panchromatic and SAR imagery~\blue{\citep{gimp2017data,howat2014greenland}}, as detailed in~\cref{app:grimp}. We use the land-ocean mask to focus the model evaluation over land only.

We also process daily 500m visible and NIR observations in 4 spectral bands from MODIS Terra~\blue{\citep{modisData}}, as they could provide key information on surface reflectance changes due to melt and other factors controlling melt, such as the deposition of particulate matter or microbial growth. However, the region experiences frequent cloud cover, limiting the availability of MODIS data.
For this reason, we do not use MODIS in our core dataset, but include it in the auxiliary dataset. 

\subsection{Data characteristics}\label{sec:data_statistics}\label{sec:data_characteristics}
The core dataset contains data from each melting season (Apr 1st -- Sep 30th) in 2017-2023, resulting in $K=529$ observed days. The non-melting season (Oct 1st -- Mar 31st) is retained for selected data streams in the auxiliary dataset. The study area measures $286.3$km $\times 163.3$km totaling $\approx46,750$km$^2$. After regridding, all input and target images have the same size $(2863\times1633 \text{pixels})$ and 100m resolution. \blue{The input channels do not contain any missing values, and }the total number of input channels is four in the core dataset (MAR, PMW, DEM, and time interpolate SAR) with an additional 22 channels in the auxiliary dataset (18x MAR, 4x MODIS). \blue{The input channel `time interpolate SAR' is a running average of SAR-derived meltwater observations excluding the target date, and is detailed in~\cref{sec:baselines} and~\cref{app:running_mean_sar}.} 
Our core dataset measures ${\sim}20$GB at float32, and we intentionally kept it small to medium-sized compared to standard superresolution datasets (see~\cref{app:data_statistics}) to minimize barriers to adoption and reuse. The auxiliary dataset is significantly larger with ${\sim}1.1$TB at float64.

\begin{figure}[t]
  \centering
  \begin{subfigure}{1.\textwidth}
      \centering
      \includegraphics[trim=1.0in 0 1.0in 0, width=0.98\columnwidth, angle = 0]{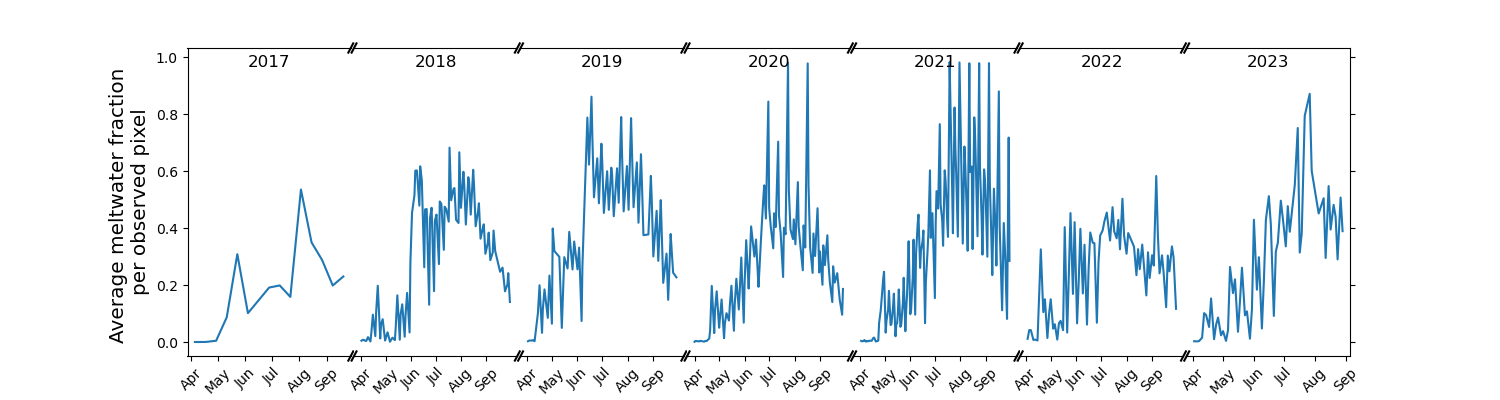}
  \end{subfigure}
  \caption[]{\textbf{Target meltwater fraction.} The plot shows the average fraction of surface meltwater per valid pixel for every retrieved SAR observation. A value of 0.7, for example, means that 70\% of the land-based study area that was observed on that day, and is not otherwise masked, contains surface meltwater. Strong fluctuations can indicate important melt events (e.g., on June 12, 2019) or noise from alternating satellite swath retrievals that cover different sections of the study area (e.g., 5x peaks from July-Sept, 2021). During 2017, our dataset only covers a subsection of the study area, and during 2022-23 S1B data is unavailable. Tick marks indicate the 1st of each month.
  }
  \label{fig:average_meltwater_per_day}
  \label{fig:data_statistics_per_day} 
\end{figure}

\Cref{fig:average_meltwater_per_day} shows the target meltwater fraction per day. The targets are represented as single-channel images with values between $0$ and $1$ that follow a bimodal distribution that is slightly skewed towards pixels with no melting (65:35), as detailed in~\cref{fig:histogram_meltwater}. The core dataset \blue{targets} contain a high percentage of \blue{missing or }invalid pixels ($63\%$), because each SAR retrieval path does not fully cover the study area (with the \% of valid pixels per image in~\cref{fig:valid_observations_per_day}), and we mask out processing artifacts and ocean-pixels ($28\%$ of the study area).

An important characteristic of our dataset is the occurrence of extreme melt events. For example,~\cref{fig:melt_event_2019_06_12} presents the data streams that were observed during the extreme melt event on June 12, 2019. During this event, Greenland's melt area almost doubled from ${\approx}30\%$ to $55\%$ within a single day (see Fig 2a in~\citeauthor{tedesco2020unprecedented},~\citeyear{tedesco2020unprecedented}).
Other extreme melt events are also visible as rapid surges in the observed meltwater fraction in~\cref{fig:average_meltwater_per_day} (e.g., 06/04/2018, 06/12/2019, 09/03/2022).
However, the satellite swath patterns also cause significant fluctuations in our observations of meltwater fraction and it becomes unclear whether some extreme melt events are not visible due to data gaps or because melting did not occur over the Helheim area (e.g., 07/31/2019, 07/28/2021, 08/14/2021, 07/18/2022).

\subsection{Test dataset and evaluation protocol}\label{sec:data_splits}
We create a validation (val) and test split that are stratified across time by randomly sampling two images from each month for each split, from 2018-2023. All leftover images from 2018-2023 are used during training, as well as 2017 during which images only cover the southwestern study area (detailed in~\cref{app:data_split}). This results in a $(13\%, 13\%, 74\%)$ split with 70 test, 70 val, and 389 training (train) days. 
Following common  protocol~\citep{goodfellow16deeplearning}, we use the train and val set for optimization of free parameters and adapting hyperparameters, respectively. All metrics and plots are reported on the test set unless noted otherwise. 

We create the stratified split \blue{due to a temporal data imbalance:~\Cref{fig:total_observations_per_month} shows that our core dataset contains some years with more observations (e.g., 2021) in comparison to others (2017). Temporally stratifying the validation and test split ensures that our evaluation metrics are equally weighted in time, rather than being biased towards the heavily sensed years. The samples that are used for validation and test are correlated to temporally nearby training samples, which may raise concerns about data leakage. However, this choice is by design as it most closely matches our task of spatiotemporally infilling data gaps. In other words, meltwater targets in the test and validation set are treated as unobserved data and models are encouraged to use all information from the training set to infer those observational gaps. We show in~\cref{app:mae_over_observation_gap} how the length of observational gaps influences model skill.} 
\strikeolive{due to the data imbalance visualized in~\cref{fig:total_observations_per_month}. Using this split, our evaluation metrics are equally weighted for any given day in any year, rather than being biased towards heavily sensed years, such as 2021. The splits also share data from the same years, such that we can evaluate the methods on creating a gap-filled product -- as opposed to forecasts. }
In summary, we choose this split to evaluate the accuracy of methods to interpolate in time and space, and we discuss future experiments towards developing methods for forecasting, reconstructing meltwater pre-SAR, and spatial extrapolation in~\cref{sec:spatial_generalization,sec:temporal_generalization}.

We avoid a cross-validation split, because the associated computational expense of repeatedly retraining models would decrease the accessibility of our proposed benchmark. \blue{Instead, we report two proxies for how sensitive our results are to resampling the splits. For each evaluation metric, we report the absolute difference between the validation and test score averaged over all models, and treat model differences smaller than this value as not significant (see test-val score difference in~\cref{tab:results_overview}). For the inferred monthly meltwater fraction, we report its spread across the train, val, and test splits (see~\cref{fig:avg_meltwater_preds_per_month_test}b).}

\subsection{Evaluation metrics}\label{sec:metrics}
\blue{Our targets contain many invalid and missing values, which complicates the computation of evaluation metrics. We address missing target values by computing most metrics as averages per valid pixel }
\strikeolive{Due to the frequent and irregular occurrence of invalid values in our targets, we compute most metrics as averages per valid pixel }- as opposed to the more commonly used averages per image. 
We apply the static land-ocean mask to all model predictions before calculating error metrics or creating plots.

Let $\tY = \{\mY_k\}_{k=0}^{K-1}$ be the set of target images, $\mY_k \in [0,1]^{(I\times J)}$, with height, $I$, width, $J$, and timestamp, $k\in\sK=\{0,...,K{-}1\}$. And, let $\hat \tY = \{\hat\mY_k\}_{k=0}^{K-1}$ with $\hat \mY_k\in [0,1]^{(I\times J)}$ be the corresponding set of predictions. Then, we start the evaluation by visualizing the predictions, $\hat \mY_k$, and biases, $\hat \mY_k - \mY_k$, for selected timestamps to gain an intuitive understanding of model skill. 

\textbf{Spatial MSE and MAE.}
We compute the spatial mean square error (MSE) per valid pixel to evaluate pixel-level prediction accuracy. Because MSE \strikeolive{is known to }encourages \blue{smoothed, }blurry results due to the quadratic penalty, we also use the spatial mean absolute error (MAE)\strikeolive{which slightly reduces this issue}~\citep{subich25doublepenalty}. The commonly used peak signal-to-noise ratio (PSNR) can be expressed as a function of MSE, and is reported in~\cref{tab:extra_metrics}. The table also contains the standard deviation of each error, the root mean square error (RMSE), and the coefficient of determination ($R^2$), although we caution against $R^2$ due to our bimodal distribution. We compute MAE and MSE with:

\begin{equation}
\text{Err}_s(\tY, \hat \tY) = \frac{1}{N_\text{valid}} \sum_{k\in\sK}\left( \sum_{(i,j)\in IJ_{\text{valid},k}}{(\lvert y_{k,i,j} - \hat y_{k,i,j}\rvert)^p}\right)
\label{eq:mse}
\end{equation}

\blue{where $\text{Err}_s$ is the mean of the p-th power of absolute pixelwise differences, pooled over all valid pixels in the dataset, and reduces to the MSE for $p=2$ and the MAE (i.e., $L_1$) for $p=1$}; $\lvert \cdot \rvert$ denotes the absolute value; $IJ_{\text{valid},k}$ is the set of all (lat, lon) index combinations, $(i,j)$, that correspond to valid pixels in the $k$-th image;
$N_\text{valid} = \sum_{k\in \sK} n_{\text{valid},k}$ is the number of valid pixels summed over all images; $n_{\text{valid},k} = |IJ_{\text{valid},k}| = \sum_{(i,j)\in IJ_{\text{valid},k}}1$ is the number of valid pixels in the $k$-th image;  $y_{k,i,j}$ and $\hat y_{k,i,j}$ are single pixels in range $[0,1]$ in the target and predicted images, respectively. In general, we denote a tensor as $\tA$, a matrix as $\mA$, and a scalar as $a$ or $A$.

\textbf{SSIM.}
Since the pixelwise MAE and MSE are less suitable for capturing the stochastic nature of our downscaling problem where multiple plausible high-resolution outputs exist for a single low-resolution input—for example along meltwater boundaries—we also compute the structural similarity index measure (SSIM). The SSIM compares two images by evaluating image statistics on a sliding window basis and ranges between $[0,1]$ with best values being $1$. To compute the SSIM on our masked images we set all invalid pixels to zero before computing the SSIM. This choice increases scores at the border to invalid pixels, but remains a fair measure for intercomparing models as the location of invalid pixels is independent of the model.

\begin{equation}
\text{SSIM}(\tY, \hat \tY) = \frac{1}{N_\text{valid}} \sum_{k\in \sK} \sum_{(i,j)\in IJ_{\text{valid},k}} \text{ssim}_\text{im}(\mM_k \odot \mY_k, \mM_k \odot \hat \mY_k)_{i,j}
\label{eq:ssim}
\end{equation}
where $\mM_k\in \mathds{1}^{(I\times J)}$ is a binary mask with $0$ for invalid and $1$ for valid pixels; $\mY_k$ is the ground-truth image tile $k$; $\hat \mY_k$ is the predicted image tile; $\odot$ denotes an element-wise product. And, $\text{ssim}_{\text{im}}\in [0,1]^{(I,J)}$ is a tensor in which each $(i,j)$-th pixel contains the value of the corresponding sliding window calculation:

\begin{equation}
\text{ssim}_\text{im}(\mY_k, \hat \mY_k)_{i,j} = 
\frac{(2\mu_{\hat y}\mu_y +c_1)(2\sigma_{\hat y y} + c_2)}
{(\mu_{\hat y}^2 + \mu_y^2 + c_1)(\sigma_{\hat y}^2 + \sigma_y^2 + c_2)}
\label{eq:ssim_im}
\end{equation}
where $\mu_y$ and $\mu_{\hat y}$ are the mean over a window centered at pixel $(i,j)$ in tile $\mY_k$ or $\hat \mY_k$, respectively (the subscripts are omitted for brevity); $\sigma_y$ and $\sigma_{\hat y}$ are the standard deviation over all pixels in the same window; $\sigma_{\hat y y}$ is the co-variance between all predicted and ground-truth values in the window; and $c_1,c_2$ are constants. The sliding window is implemented using a Gaussian kernel with size $h_w=72$,
standard deviation $\sigma_w=10$,  reflection padding, and constants 
$c_1=1e{-}4$
and $c_2=9e{-}4$.
The sliding window implementation is based on the torchmetrics package~\citep{detlefsen22torchmetrics}.

\textbf{Binary classification metrics.} 
For interpretability, we also compute classification metrics by categorizing predicted and target surface meltwater fraction into `no-melt' and `melt' using a threshold of $y_\text{thresh} = 0.1$. We compute precision ($\frac{tp}{tp+fp}$), recall ($\frac{tp}{tp+fn}$), and accuracy ($\frac{tp+tn}{tp + tn + fp + fn}$), with the abbreviations $t$=true, $f$=false, $p$=positive, and $n$=negative. To account for the invalid data masks, we compute the scores as averages across all valid pixels, for example, for accuracy:

\begin{equation}
\text{Acc}(\tY, \hat \tY) = \frac{1}{N_\text{valid}} \sum_{k\in \sK} 
\sum_{(i,j)\in IJ_{\text{valid},k}}
{\mathds{1}\left[\mathds{1}(y_{k,i,j} > y_\text{thresh}) == \mathds{1}(\hat y_{k,i,j} > y_\text{thresh})\right]}
\label{eq:accuracy}
\end{equation}

We also compute an F1-score which is the harmonic mean of precision and recall. The equation for each metric is given in~\cref{app:metrics}.

\textbf{Number of parameters.} We count the number of parameters per model to provide a measure of model complexity. We count the number of weights (i.e., free parameters) and hyperparameters that are adjusted using our training and/or validation dataset and exclude fixed hyperparameters, free parameters fitted using external datasets, and data generating parameters, such as MAR parametrizations or sensor calibrations.

\textbf{Monthly statistics.} We evaluate if a model systematically under- or overestimates meltwater across the study area by calculating the monthly-averaged meltwater fraction per valid pixel:

\begin{equation}
\overline{\overline{\hat y}}_m = \frac{1}{\lvert \sK_m \rvert} \sum_{k\in \sK_{m}} \frac{1}{n_{\text{valid},k} } \sum_{(i,j)\in IJ_{\text{valid},k}} \hat y_{k,i,j}
\label{eq:monthly_meltwater_fraction}
\end{equation}

where ${\lvert \sK_m \rvert}$ is the number of images in month, $m$, and all years and $\sK_m$ indexes those images.

\subsection{Downscaling methods}\label{sec:models}

\blue{We implement several methods that are inspired by existing practice, including a random forest and methods that rely on thresholding or interpolating a single data stream. We refer to them as traditional methods and compare them with UNet-based deep learning architectures.}
\strikeolive{We implement several traditional methods that are used in practice and compare them with UNet-based deep learning architectures.}

\subsubsection{Traditional methods}\label{sec:baselines}
\textbf{Time interpolate SAR.}\label{sec:running_mean_sar}
The time interpolate SAR model assumes that the meltwater fraction at any unobserved target date is approximately the running monthly mean of all observations surrounding the target date, and is inspired by~\citep{wang2007melt,li2024glacier} and detailed in~\cref{app:running_mean_sar}. This model uses only imagery from the training set as inputs to avoid data leakage during validation or test. \blue{The target image itself is also excluded from the running mean. }The predictions of this model are also used as inputs to the deep learning-based models in most of our experiments.

\textbf{Interpolate MAR.}\label{sec:interpolate_mar}
\blue{The interpolate MAR model assumes that the meltwater fraction is a monotonic function of the MAR-modeled average liquid water content within the top meter of the snowpack. Whereas previous studies apply a hard threshold to this variable for a binary melt estimate~\citep{fettweis2011melting,datta2018melting,dethinne2023sensitivity}, we apply a smooth threshold, tuned on the validation set, to retrieve a continuous meltwater fraction, as detailed in~\cref{app:interpolate_mar}. }
\strikeolive{The interpolate MAR model assumes that the meltwater fraction equals the MAR-modeled average liquid water content within the top meter of snow after applying a Gaussian blur and adjusting the intensity, as detailed in~\cref{app:interpolate_mar}. }We interpret this model as the surface meltwater fraction predicted by the MAR model.

\textbf{Threshold PMW.}\label{sec:threshold_pmw}
The threshold PMW model is binary and detects meltwater if the brightness temperature at a given location exceeds a threshold that is based on the winter mean brightness temperature and a fixed output of a microwave emission model of layered snowpack. This method is commonly used in large-scale studies across \strikeolive{Greenland} \blue{the cryosphere~\citep{abdalati1995passive,liu2006spatiotemporal,tedesco2007snowmelt,tedesco2009assessment,semmens2013recent} and our implementation follows a recent study on high-resolution melt estimates over the Greenland ice sheet~\citep{colosio2020surface}}, as detailed in~\cref{app:threshold_pmw}. 

\textbf{Threshold DEM.}\label{sec:threshold_dem}
The threshold DEM model assumes that all locations within a monthly varying elevation band contain meltwater, as detailed in~\cref{app:threshold_dem}. This follows the intuition that higher altitudes will exhibit colder temperatures and lower altitudes are more likely to feature exposed rock, registering less melt, while the intermediate area may be melting. \blue{While the exact formulation is specific to our work, it draws on previous work that uses elevation to downscale or interpolate melt-related data from models \citep{noel2016daily,noel2023higher,tedesco23maranddem,lipscomb2013implementation,fischer2014system} and observations \citep{scher2021mapping,reeh1991parameterization}.}

\blue{
\textbf{Random Forest.}\label{sec:random_forest}
Our random forest model uses the same inputs as the UNet but is trained pixelwise, thereby assuming spatial independence between pixels. Our implementation is guided by~\cite{irvin21fluxnet} and detailed in~\cref{app:random_forest}. Random forests are a common choice in remote sensing and can outperform neural networks depending on the dataset~\citep{yuval20randomforests,irvin21fluxnet}.
}

\subsubsection{Deep learning-based downscaling methods}\label{sec:deeplearning}
\textbf{UNet.}\label{sec:unet}
We implemented a UNet-based model~\citep{ronneberger15unet}, because it is a commonly used model architecture to approach image-to-image translation problems using an \blue{$L_1$ or MSE}~\citep{sha20unetl1}, adversarial~\citep{isola17pix2pix}, or diffusion-based~\citep{saharia23sr3} loss function, and continues to achieve competitive results\strikeolive{ on medium-sized datasets~\citep{isensee21nnunet}}\blue{~\citep{cachay26ucast}}. The UNet architecture is designed to learn small and large-scale spatial correlations by using an encoder-decoder architecture with skip connections, as visualized for a vanilla UNet in~\cref{fig:unet_architecture}. Further, using convolutional layers introduces a theoretically beneficial spatial locality bias~\citep{cachay21gnns}. The UNet learns the mapping, $f_\theta:\tX_k \rightarrow \mY_k$, with the inputs, $\tX_k \in \mathds{R}^{n_c\times w \times h}$, that have $n_c=4$ input channels (MAR WA1, PMW, DEM, time interpolate SAR) and outputs, $\mY_k \in \mathds{R}^{w \times h}$, that is the surface meltwater fraction, as displayed in~\cref{fig:overview}. Instead of using the full study area in the in- and outputs, we train the UNet model on fixed-size tiles with size $(w\times h)$. We train the model to minimize \blue{an average L$_1$-loss across valid target pixels }\strikeolive{a masked L$_1$-loss }, given by~\cref{eq:mse} with $p=1$ and $K$ equal to the batch size. 

To train the UNet, we implemented a dataloader that randomly generates locations from which to extract the tiles from within the study area. These tile locations are resampled for every image and epoch, whereas one epoch contains one tile of every image in the dataset. In comparison with the commonly used approach of precomputing and storing tiles on disk~\citep{stengel20phiregan}, this dynamic tiling approach reduces redundant storage and, more importantly, can be seen as a form of data augmentation that encourages translation equivariance. Memory consumption does not become an issue due to GeoTIFFs supporting windowed reading. During validation and test, we create a full-size image mosaic by \blue{sliding the entire learned model across the study area in a sliding-window fashion }with a stride, $s$, and merging predictions by computing the average value at every overlapping pixel. We erode the outer $e$ pixels \blue{of each tile }before mosaicing, because the accuracy of CNN-based predictions is known to deteriorate towards \blue{tile }\strikeolive{image }borders.
For tuning hyperparameters, we compute all evaluation metrics on the mosaic and use SSIM to select the best model configuration. 

\textbf{Model optimization.}\label{sec:deeplabv3plus} We distinguish our experiments to optimize the UNet as: `vanilla UNet', `DeepLabv3+', and `UNet SMP'. The vanilla UNet follows~\citep{ronneberger15unet}, for the most part and is detailed in~\cref{app:unet_vanilla}. Using the vanilla UNet, we ran early experiments on the choice of loss function, $L_1$ vs MSE, tile size, $[64,128,256,512]$, and preprocessing steps, and observed the best SSIM scores with an $L_1$-loss, tile size $h=w=512$, and a Gaussian blur that smooths MAR and PMW inputs. The vanilla UNet is comparatively shallow, trained from scratch, and has a receptive field that limits the maximum learnable spatial correlations to $188$px, as detailed in~\cref{app:unet_vanilla}. Thus, we follow~\citep{chen18deeplabv3plus} using the segmentation models pytorch (SMP) library to implement the more complex UNet SMP and DeepLabv3+\strikeolive{, which both use a deeper encoder with pretrained weights}. 
\blue{The UNet SMP mainly differs by replacing the shallow non-pretrained encoder with a deeper, ImageNet-pretrained encoder, which enlarges the receptive field (see~\cref{app:unet_smp}). }DeepLabv3+ \blue{also uses a pretrained encoder and }adds a special module in the bottleneck that extracts multi-scale features using dilated convolutions (called atrous spatial pyramid pooling) but limits the reference implementation to $h=w=304$px (see~\cref{app:deeplabv3plus}). Thus, we still optimize the commonly used hyperparameters in DeepLabv3+, but focus further experiments on UNet SMP. The final model, UNet SMP from now on referred to as `UNet', uses $h=w=512$, $s=480$, $e=16$, ImageNet-pretrained weights, and an xception71 encoder, with the full set of parameters detailed in~\cref{app:unet_smp}.

\section{Results}\label{sec:results}

In~\cref{sec:results_spatial}, we demonstrate that the UNet and time interpolate SAR models resolve surface meltwater at higher spatial resolution and evaluation scores than the threshold PMW and interpolate MAR models.
In~\cref{sec:results_time}, we show that the PMW- and MAR-based models over- or underestimate meltwater during selected months with respect to the SAR-derived targets, whereas the UNet and time interpolate SAR model can capture the seasonal meltwater variation more accurately. \blue{In~\cref{sec:feature_importance}, we demonstrate that time interpolate SAR is the UNet's most important, but not only, informative feature.} In~\cref{sec:results_inference}, we analyse the generated product for every day in the melting seasons of 2017-2023. \blue{In~\cref{sec:station_obs}, we compare to station observations.} In~\cref{app:results}, we also report on the difference between the deep learning-based models.

\subsection{High-resolution features in model predictions}\label{sec:results_spatial}

\begin{figure*}[t]
  \centering
      \includegraphics [width=0.99\textwidth, angle = 0]{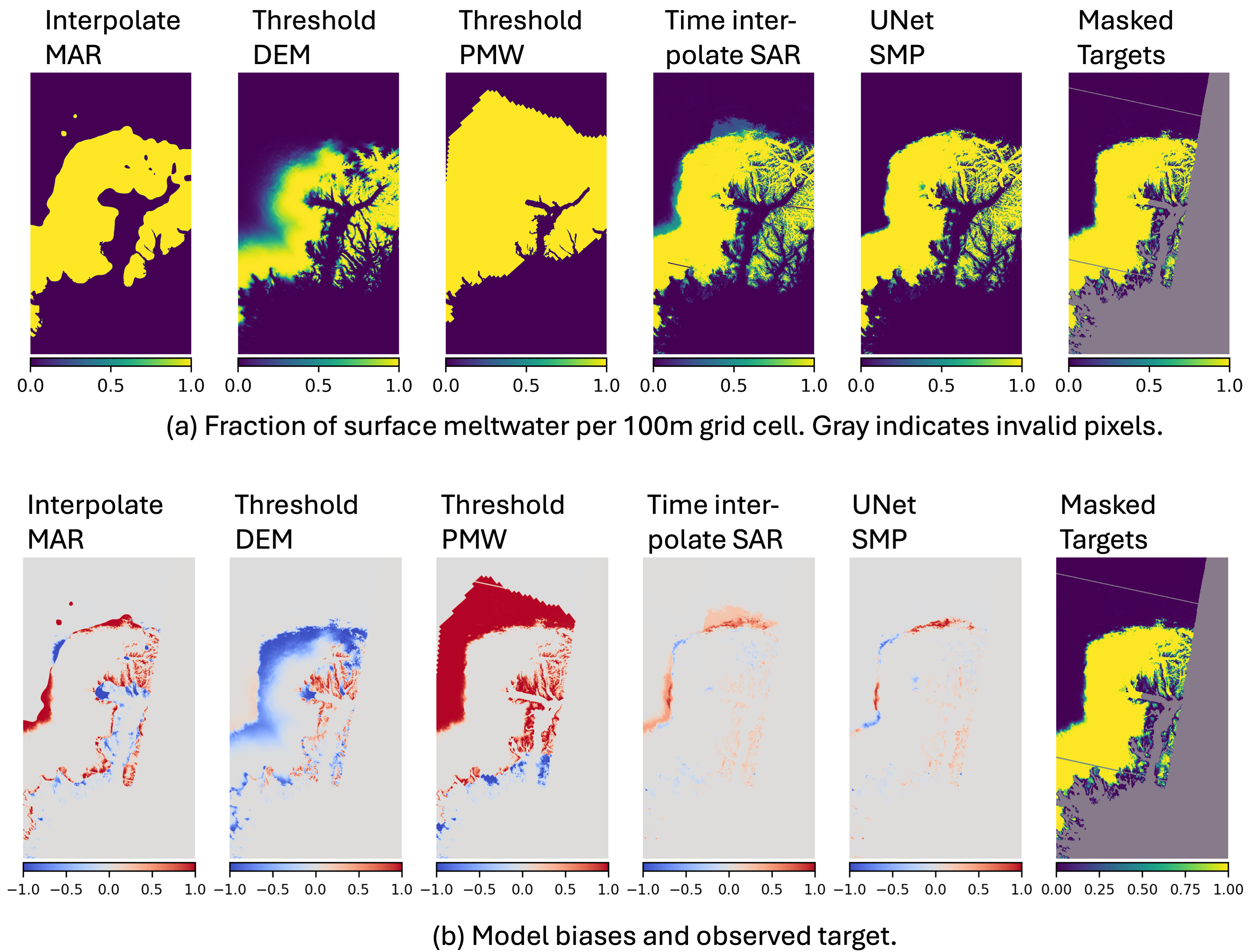}
\caption[Model bias sample]{\textbf{High-resolution predictions and biases.} Prediction (top) and bias (bottom) of each model (left 5 columns) and observed meltwater target (rightmost column) for a date in the test dataset (June 24, 2018). Model biases are calculated as prediction minus target, such that red and blue indicate over- and underestimated values, respectively. The input data streams for this date are displayed in~\cref{fig:overview}, and the date was selected to highlight each model's characteristics.
}
\label{fig:model_vs_targets_2018_06_24} 
\end{figure*}

~\Cref{fig:model_vs_targets_2018_06_24}a and~\cref{fig:model_vs_targets_2018_06_24}b show a map of each model's surface meltwater prediction and bias for a selected date in June, 2018. Every model's predictions have the same high-level features: The models generate predictions over the date's unobserved pixels (gray area), correctly predict a region of no-meltwater in high elevations (Northern deep blue area), and predict significant melting at medium-to-low elevations (contiguous yellow area) toward the ocean.
Predicting high-resolution details along meltwater boundaries and the mountainous coast seems more challenging: The interpolate MAR and threshold PMW predictions are too coarse and limited by their native resolution of 3-5km.
The time interpolate SAR prediction resolves high-resolution features at $100$m$/$px, but contains postprocessing artifacts (blue lines or sharp green-to-blue transitions).
The UNet model resolves high-resolution features with similar biases as time interpolate SAR, e.g., along the Northern meltwater boundary, but is able to partially correct for postprocessing artifacts.

\label{sec:results_table}
\Cref{tab:results_overview} shows that the deep learning models (vanilla UNet, UNet, and DeepLabv3+) and time interpolate SAR achieve the best evaluation scores across our study area and observations in the test dataset. Most of the scores are computed on a per-pixel basis, meaning that high scores are likely to be associated with a model's capability for predicting high-resolution features. Notably, the UNet achieves an accuracy of $95\%$, which significantly improves on the operational threshold PMW model ($72\%$) and a na\"ive model that would always predict no-melt ($65\%$).
\blue{The UNet also improves over the random forest ($91\%$), possibly owing to its larger receptive field or capacity for fitting nonlinear functions.} 
The UNet outperforms time interpolate SAR in terms of error metrics by achieving a ${\sim}40\%$ lower MAE and predicting fewer false positives (as indicated by the improved precision at comparable recall in~\cref{tab:extra_metrics}).
However, the UNet model has 49.6M free parameters, which is significantly more than time interpolate SAR, and can be an important evaluation criterion (see~\cref{sec:limitations}). 

\begin{table}[t]
  \caption{\textbf{Results table.} Evaluation scores on test dataset by model, with \#p denoting the number of parameters. The last row reports the average difference between the models' val and test scores (see~\cref{sec:data_splits}). Best scores within the test-val score difference are \textbf{bold}, next-best \textit{italic}, reported to three significant digits, with standard deviations in~\cref{tab:extra_metrics}. `UNet SMP' is abbreviated as `UNet'.
}
  \centering
  \resizebox{0.99\textwidth}{!}{
    \begin{tabular}{lllllll}
      \toprule
      Model     &  \#p. $\downarrow$ &MAE$_s$ $\downarrow$&MSE$_s$ $\downarrow$&  Acc. $\uparrow$&F1 $\uparrow$& SSIM$_{\sigma=10}$$\uparrow$\\
      \midrule
      Time interpolate SAR & \textbf{1} &0.0778&0.0389&  0.899
   &0.812& 0.711\\
      Interpolate MAR &  \textit{4}&0.167&0.149&  0.826
   &0.664& 0.493\\
   Threshold PMW&  \textbf{0}&0.272& 0.254&  0.724
   &0.557& 0.423\\
      Threshold DEM&  12&0.187&0.137&  0.813
   &0.699& 0.462\\
      \blue{Random Forest} &  \blue{5.94M}&\blue{0.0732}&\blue{\textit{0.0303}}&  \blue{0.908}
   &\blue{0.814}& \blue{0.692}\\
      DeepLabv3+      &  42.9M&\textit{0.0572}&\textit{0.0331}&  \textit{0.935} & \textit{0.830} & 0.711\\
      UNet&  49.6M&\textbf{0.0474}&\textbf{0.0250}&  \textbf{0.946} &\textbf{0.848}& \textbf{0.762}\\
      vanilla UNet&  31.0M&\textit{0.0594}&0.0356&  \textit{0.934} &\textbf{0.840}& \textit{0.744}\\
      \hline
      test-val score difference  &  n/a&0.00475&0.00474&  0.00480 &0.0120& 0.00574\\
      \bottomrule
    \end{tabular}}
  \label{tab:results_overview}
\end{table}

\subsection{Temporal biases in meltwater predictions}\label{sec:results_time}
\begin{figure}[t]
  \centering
  \includegraphics[width=0.99\linewidth, angle = 0]{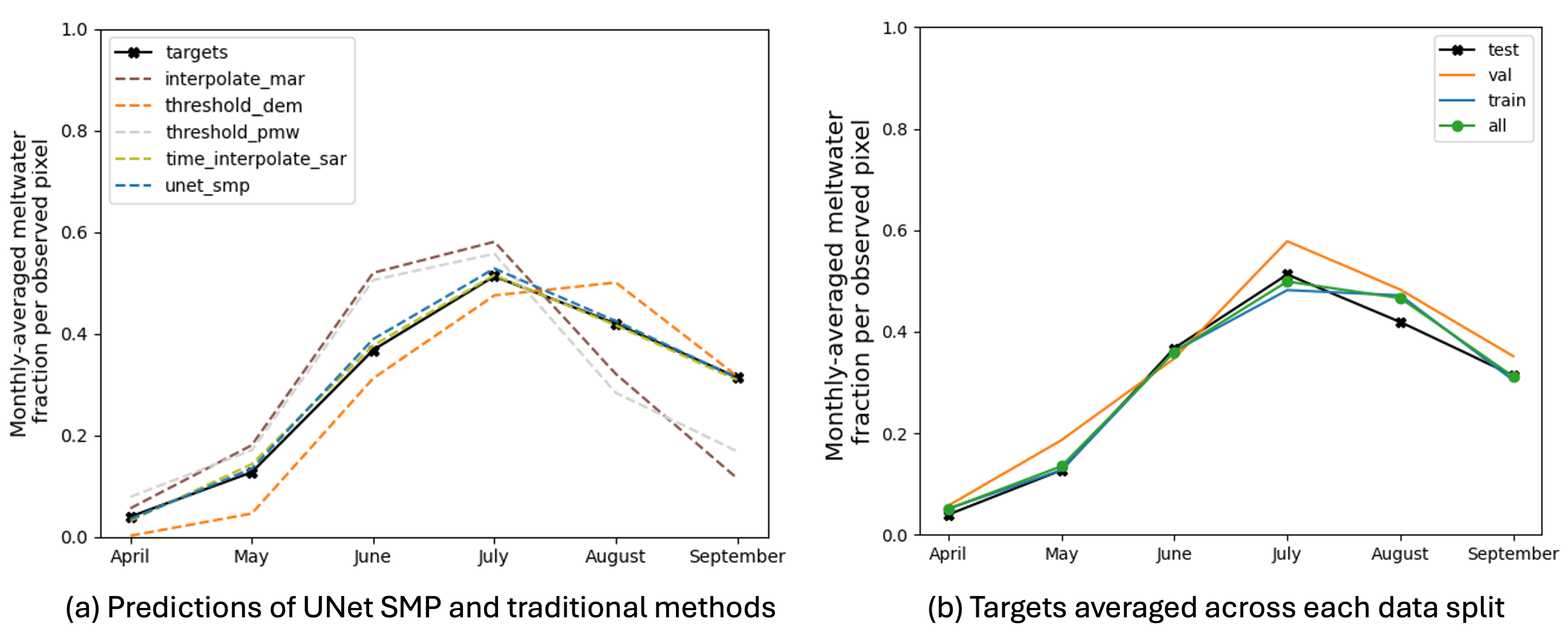}
\caption[]{\textbf{Average meltwater fraction over a melting season.} Fig. 6a) shows the target (black) and predicted monthly-averaged surface meltwater fraction per valid pixel, averaged across 2017-2023. \blue{Fig. b) shows the targets only (no predictions) -- averaged across the test (black), val (orange), train (blue), and entire (green) dataset -- to illustrate the variation of monthly-averaged surface meltwater fraction across different data splits.}
\strikeolive{Fig. 6b) shows the same quantity but averaged across the train, val, test, and entire dataset to illustrate the magnitude of variation between data splits.}}
\label{fig:avg_meltwater_preds_per_month_test} 
\end{figure}

To uncover temporal biases, we plot the predicted and observed surface meltwater fraction per month in~\cref{fig:avg_meltwater_preds_per_month_test}a.
Notably, interpolate MAR (brown) and threshold PMW (gray) both overpredict meltwater in early summer (June) and underpredict it in late summer (Sept) with respect to the SAR targets. 
The UNet (blue) and time interpolate SAR (green) more accurately match the SAR observations (solid black) in terms of the average monthly surface meltwater fraction across the study period. Other differences across model predictions (e.g., in \strikeolive{June} July) are not interpreted, 
because they are smaller than the variation between data splits \blue{of the same month's target meltwater fraction} in \cref{fig:avg_meltwater_preds_per_month_test}b, and would not necessarily persist on resampled data splits.

\blue{As the models are tasked to fill in spatiotemporal gaps in observations over shorter (1-3 day) and longer (4-12 day) distances, we also plot the UNet MAE over the length of this observational gap:~\Cref{fig:mae_over_observation_gap} illustrates that the MAE worsens with longer observational gaps but remains acceptable (MAE$\approx$0.055 for 6+ days).}

\subsection{Feature importance}\label{sec:feature_importance}
\blue{To analyze the sensitivity of the UNet with respect to each input channel, we retrain the UNet multiple times using the same hyperparameters but withholding or exclusively using one of the input channels, similar to~\cite{husman24meltwater}. The evaluation scores in~\cref{tab:feature_importance_metrics} suggest that time interpolate SAR is the most informative input channel, i.e., causing the largest drop in scores when being withheld and achieving the best scores when using only one channel. The UNet scores are also sensitive to removing the MAR channel, and are not significantly impacted by removing the DEM or PMW inputs.~\Cref{fig:avg_meltwater_preds_per_day_test_only_sar} shows that a UNet trained exclusively on time interpolate SAR inputs underpredicts the test-set meltwater fraction on June 28th and July 12th, 2023, and that adding the other channels recovers this drop. Overall, this suggests that the MAR channel may be important during rapidly changing periods, such as the July 2023 heat event, when the running mean calculation in time interpolate SAR is too smooth. 

}

\begin{table}[t]
  \caption{\blue{\textbf{Feature importance.} Evaluation scores on the test dataset for a UNet SMP that was retrained each time withholding or exclusively using one of the input channels. Best scores within the test-val score difference from~\cref{tab:results_overview} are bold, next-best italic.}
}
  \centering
  \resizebox{0.82\textwidth}{!}{
  \begin{tabular}{llllll}
    \toprule
    Model     & MAE$_s$ $\downarrow$&MSE$_s$ $\downarrow$&  Acc. $\uparrow$&F1 $\uparrow$& SSIM$_{\sigma=10}$$\uparrow$\\
    \midrule
    UNet all channels&\textbf{0.0474}&\textbf{0.0250}&  \textbf{0.946} &\textbf{0.848}& \textbf{0.762}\\
    UNet no DEM        & \textbf{0.0470}&\textbf{0.0250}&  \textbf{0.947} &\textbf{0.849}& \textbf{0.760}\\
    UNet no PMW        & \textbf{0.0457}&\textbf{0.0242}&  \textbf{0.949} &\textbf{0.851}& \textbf{0.762}\\
    UNet no MAR        & \textit{0.0586}&\textit{0.0361}&  \textit{0.935} &\textit{0.832}& \textit{0.744}\\
    UNet no SAR        & 0.0677&0.0427&  0.924 &0.792& 0.673\\
    UNet only DEM      & 0.2294&0.1972&  0.753 &0.629& 0.471\\
    UNet only PMW      & 0.1977&0.1629&  0.792 &0.614& 0.446\\
    UNet only MAR      & 0.1087&0.0808&  0.881 &0.728& 0.551\\
    UNet only SAR      & \textit{0.0606}&\textit{0.0379}&  \textit{0.932} &\textit{0.829}& \textit{0.744}\\
    \hline
    test-val score difference  &0.00475&0.00474&  0.00480 &0.0120& 0.00574\\
    \bottomrule
  \end{tabular}}
  \label{tab:feature_importance_metrics}
\end{table}

\subsection{Gap-filled surface meltwater product}\label{sec:results_inference}

\begin{figure*}[t]
  \centering
  \begin{subfigure}{1.\textwidth}
    \centering
    \includegraphics [trim=0.5in 0 1.1in 0, clip, width=0.99\textwidth, angle = 0]{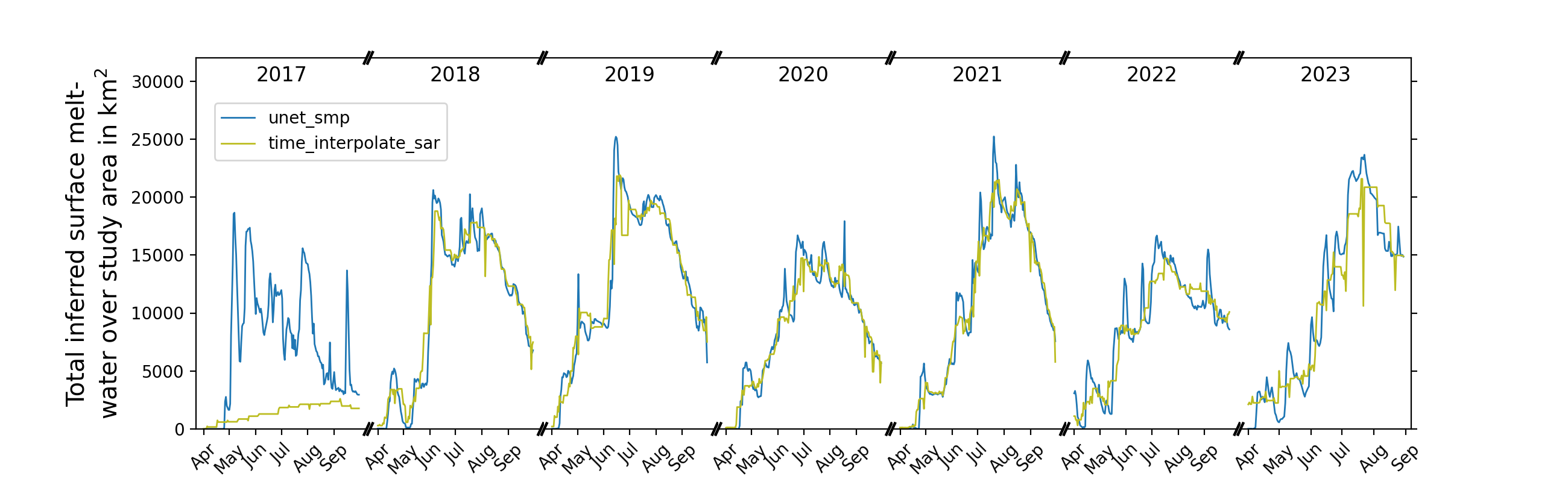}
    \caption{UNet and time interpolate SAR}
    \label{fig:meltwater_extent_per_day_unet}
  \end{subfigure}
  \centering
  \begin{subfigure}{1.\textwidth}
    \centering
    \includegraphics [trim=0.5in 0 1.1in 0, clip, width=0.99\textwidth, angle = 0]{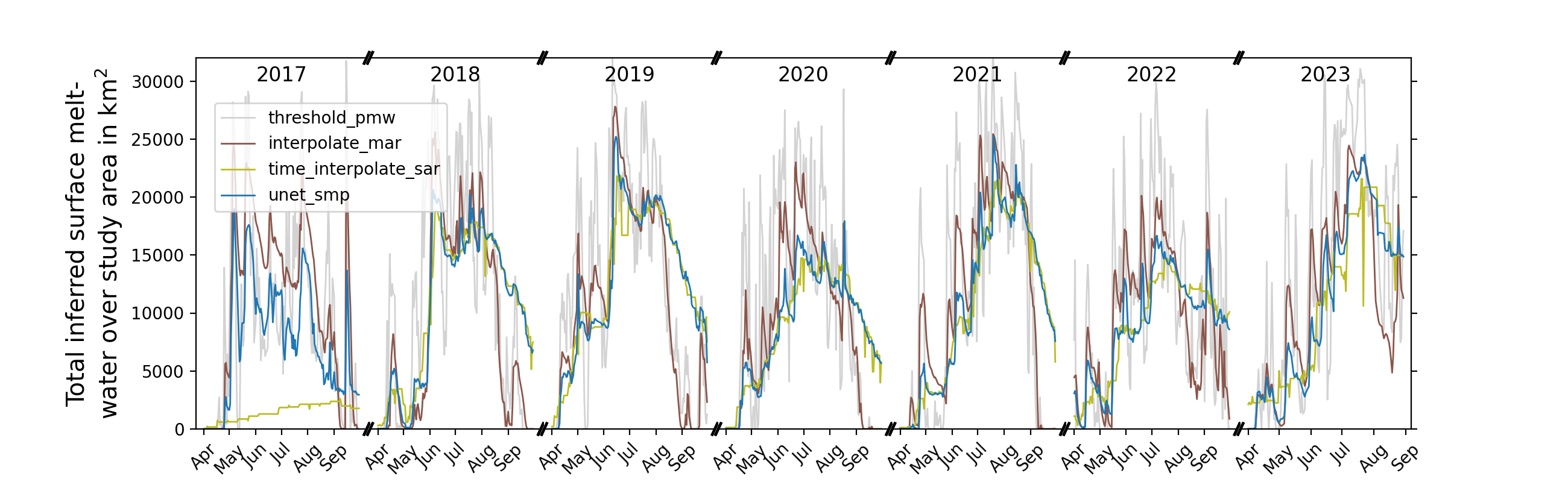}
    \caption{UNet, threshold PMW,  interpolate MAR, and time interpolate SAR}
    \label{fig:meltwater_extent_per_day_traditional}
  \end{subfigure}
\caption[]{\textbf{Predictions of surface meltwater for every day in the 2017-2023 study period.} Predicted total area of surface meltwater within the study area for the UNet (blue), time interpolate SAR (yellow), threshold PMW (gray), and interpolate MAR (brown) model. Vanilla UNet and DeepLabv3+ are reported in~\cref{fig:meltwater_extent_per_day_unet_deeplabv3}. Sharp increases in total meltwater, such as on June 12, 2019, indicate extreme melt events. The target observations are not plotted due to their spatiotemporal gaps. Tick marks indicate the 1st of each month.
} 
\label{fig:meltwater_extent_per_day} 
\end{figure*}

After training, we query each model to predict a map of surface meltwater for every day in the study period.~\Cref{fig:meltwater_extent_per_day} shows the area that lies within the study area and is predicted to be covered by meltwater. In this plot there are no targets because the SAR retrieval paths only intersect with subsections of the study area. But, for completeness we plot the predicted meltwater fraction per observed pixel against the comparatively noisy test set targets in~\cref{fig:avg_meltwater_preds_per_day_test}.

~\Cref{fig:meltwater_extent_per_day_unet} shows that extreme melt events are visible in the UNet-predicted gap-filled meltwater product as sharp increases, e.g., during the event on June 12, 2019. Moreover, some extreme melt events, e.g., Aug. 14, 2021~\citep{moon21extrememelt} are challenging to distinguish from noise in the sporadic SAR observations (\cref{fig:average_meltwater_per_day}) and the PMW-based product (\cref{fig:meltwater_extent_per_day_traditional}), whereas their impact on meltwater surrounding the Helheim Glacier is visible in the UNet and MAR-based products in~\cref{fig:meltwater_extent_per_day_traditional}.

The time interpolate SAR model predicts less meltwater than the UNet on extreme melt days, e.g., on June 12, 2019, which is likely due to the method's moving window average. There are also step function artifacts in the time interpolate SAR predictions, e.g., in mid-June 2019, but they could likely be removed by using a larger window average. In 2017, time interpolate SAR predicts significantly less meltwater coverage than the other methods, which is due to the SAR targets over the Eastern part of our study area being masked throughout the whole year.

A video demonstrates the UNet-created product of daily 100m maps from 2018-2023, and is accessible in the supplementary material and at \href{https://youtu.be/OaonUT6dIbg}{this link}. The video illustrates the seasonal changes in surface meltwater distribution and effects of the coastlines and topography. The visible high-resolution dynamics include the redistribution of meltwater towards the Helheim glacier's border in the late season. The video also demonstrates some artifacts from the UNet's prediction stride and SAR processing artifacts.

\subsection{Station observation analysis}\label{sec:station_obs}
\begin{figure}[t]
  \centering
  \includegraphics[width=0.99\linewidth, angle = 0]{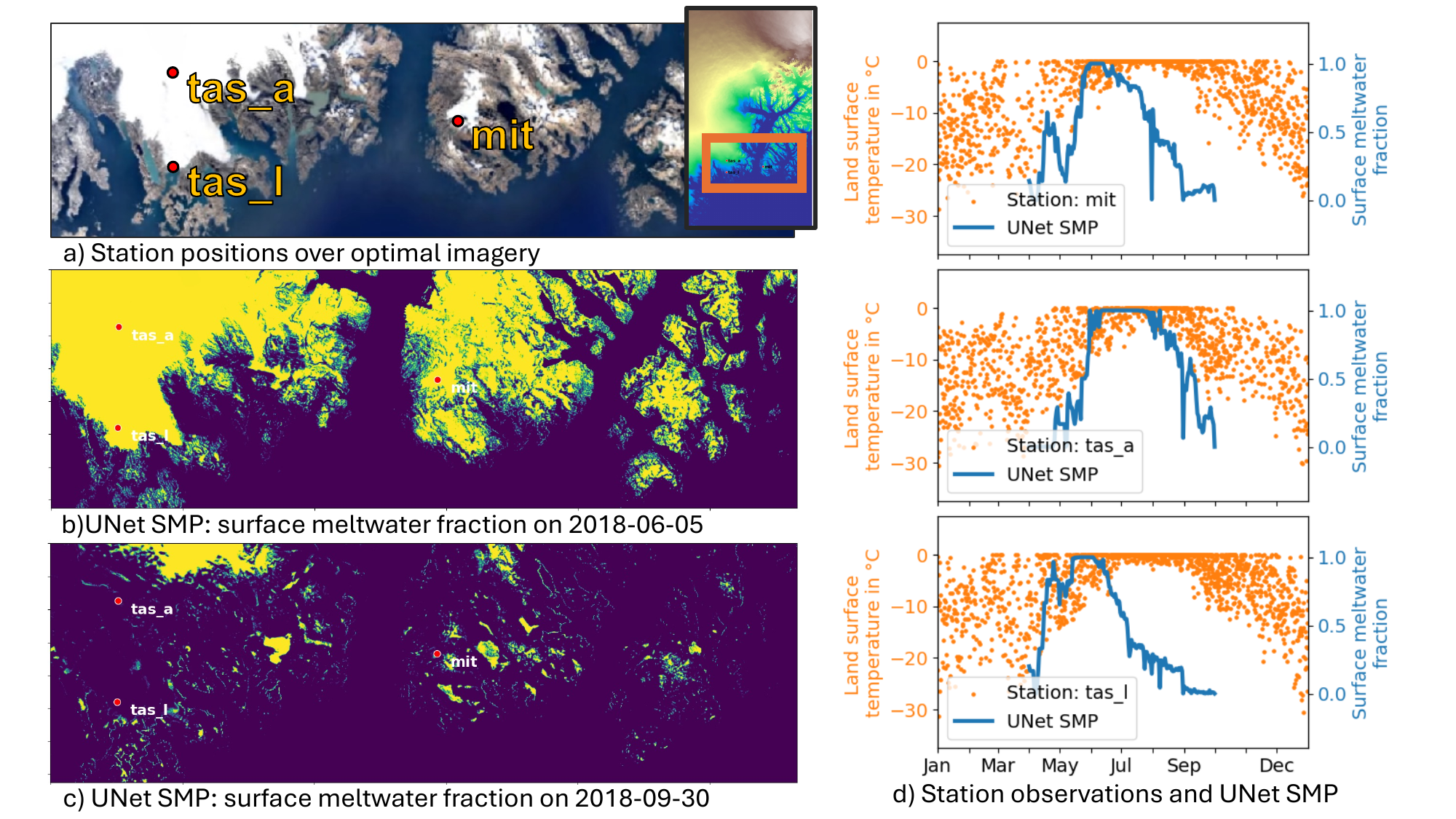}
\caption[]{\blue{\textbf{Station observation analysis.}~\Cref{fig:meltwaterbench_station_analysis}a) shows the position of the PROMICE weather stations over a 2017 optical imagery mosaic from Google Earth.~\Cref{fig:meltwaterbench_station_analysis}b,c) show the UNet SMP predicted surface meltwater fraction (yellow=melt) on June 5th and Sept. 30th, 2018.~\Cref{fig:meltwaterbench_station_analysis}d) shows all 6am local time land surface temperature observations for each station, pooled across 2018 to 2023 (orange), and UNet SMP predicted surface meltwater fraction at each station location, averaged across 2018 to 2023 (blue). X-ticks indicate the 1st of each month and~\cref{fig:meltwaterbench_station_analysis_complete} shows the same data without temporal aggregation.}}
\label{fig:meltwaterbench_station_analysis} 
\end{figure}

\blue{
~\Cref{fig:meltwaterbench_station_analysis} compares the UNet SMP against PROMICE weather station observations~\cite{fausto21promice}. Three stations lie within our study area (mit, tas-a, tas-l). Because the stations do not directly observe surface meltwater, we compare against land surface temperature (LST), under the common assumption that surface melting occurs when station LST$\geq-1^\circ$C~\cite{fausto16energyfluxes,zheng22meltflux,fang23lst}. The station records contain many gaps, so we focus on a climatological comparison by plotting all valid observations on a single-year time axis. For the UNet SMP, we plot the predicted surface meltwater fraction at the spatially corresponding pixel, averaged across 2018 to 2023. 

During the early melting season (April--June), the aggregate UNet SMP and station observation statistics agree: both increase over time, with occasional melt in April to mid-May and near-continuous melt from mid-May to mid-June. In the late season (July--October), UNet and observations show similar trends over the tas-a station but disagree at the mit and tas-l stations. Plotting the individual station observations without temporal aggregation in\cref{fig:meltwaterbench_station_analysis_complete} further confirms these trends. The late-season disagreement may arise from coastal biases (tas-a is further inland) or fresh snowfall influencing the SAR-derived targets. The stations tas-l and mit may also be sitting on exposed rock late in the season, which could explain high land surface temperatures without melting (\cref{fig:meltwaterbench_station_analysis}a). 
}

\subsection{Deep learning hyperparameter insights}\label{sec:hyperparameters}
Tuning the hyperparameters of the deep learning models required a sizable amount of computations, and increased the scores, e.g., of the UNet from an off-the-shelf performance of $0.715$ to $0.762$ SSIM. Here, we report our insights on the optimal hyperparameters in anecdotal fashion because a full documentation is beyond the scope of this paper. Notably, increasing the tile size, in particular from 64 to 256px, improved the vanilla UNet performance, which indicates a capability of the deep learning models to correct large-scale biases. Further, using ImageNet-pretrained instead of randomly initialized weights slightly increased scores ($0.01-0.02$ SSIM) across models (UNet, DeepLabv3+) and backbones choices (xception, resnet), which may be surprising due to the substantial domain shift between natural and geospatial imagery in ImageNet and MeltwaterBench~\citep{rolf24modality}. Choosing deeper encoders (xception-41, -65, -71) also improved SSIM evaluation scores. One training run of the UNet on 1x V100 GPU takes 16.5 hours \blue{including per-epoch evaluation}, with 1085 epochs, \strikeolive{500 }\blue{389} tiles/epoch, batch size $8$, $512$px tiles, and $100$m/px. Inference time is significantly faster, \blue{at 20 tiles/second on one V100 GPU and 18 CPU cores -- it would take $\approx$2.5 hours to create predictions for all days in a melting season (Apr-Sept) across all of Greenland ($\sim 2.2$M km$^2$), assuming linear scaling of compute by processed area, batch size $8$, $480$px stride, $16$px erode size.} \strikeolive{Running inference across all of Greenland ($\sim 2$M km$^2$) would take on the order of tens of minutes to hours.} Additional hyperparameter insights are detailed in~\cref{app:hyperparameters}.

\section{Discussion}\label{sec:discussion}

\subsection{Implications of seasonal bias for process-based models}\label{sec:discuss_mar}
\label{sec:discuss_baselines}
Our results in \cref{fig:avg_meltwater_preds_per_month_test}a have shown that current MAR- and PMW-based methods have a systematic seasonal bias in estimating the Helheim glacier's distribution of surface meltwater, in comparison to SAR-derived targets. 
These biases could stem from differences in the instruments: SAR might be more sensitive than PMW to deep liquid water accumulation in the late season, whereas PMW may be more sensitive to early season shallow meltwater. Further, MAR is consistent with PMW similar to past studies that have also analysed MAR outputs over the top 1~m of snow~\citep{dethinne2023sensitivity,datta2018melting,fettweis2011melting}.

Additionally, the seasonal biases may result from processes that are not sufficiently captured in MAR: The underestimation of meltwater in late summer by MAR, in particular, might be due to light-absorbing impurities, which are not accounted for in the model, leading to an overestimation of albedo~\citep{antwerpen2022assessing}. Another reason might be the so-called ``mixed pixel effect'': for example, if a 3-25km PMW pixel contains a mixture of rock and snow, and the rock heats up faster during early summer, it can skew the pixel's average over the melt-threshold despite the snow not yet having melted. This would imply that km-scale data from PMW or MAR can exhibit biases over a smaller spatial domain that could also affect other sites in Greenland. Our analysis in \cref{fig:avg_meltwater_preds_per_month_test}a has shown that the UNet is capable of accounting for these seasonal biases and correcting them -- creating a product that more closely resembles SAR.

Another reason for the seasonal bias may be our data selection: Our benchmark contains SAR data from the morning hours, PMW data from the evening, and daily averages from MAR. Some of the over- (under-) estimation of PMW and MAR during the peak (late) melting season may be attributable to melting (snowfall) that occurs throughout the day. We could have chosen the PMW morning pass to allow for a closer comparison between the SAR and PMW sensors. But, PMW and SAR observations are not guaranteed to be temporally aligned in other areas, and our experiments on the evening pass demonstrate that the UNet can account for the systematic bias introduced by this commonly occurring temporal mismatch.

\subsection{Limitations of the benchmark and daily gap-filled product}\label{sec:limitations}
The data-driven downscaling models described in this paper are fundamentally constrained by the accuracy of the \strikeolive{``ground-truth''} surface meltwater targets, which we derive from SAR HH-polarization backscatter intensity using a threshold-based approach. \blue{While} SAR is generally considered one of the \strikeolive{most accurate} \blue{best-suited} methodologies for inferring surface meltwater presence due to its high resolution (10m), sensitivity to liquid water, and indifference to cloud presence \blue{\origcitep[e.g.][]{johnson2020evaluation}}, \strikeolive{. But,} SAR-derived meltwater estimates\blue{, like any remote sensing product,} \strikeolive{can} contain \strikeolive{inaccuracies} \blue{inherent uncertainties} due to local variations in the optimal threshold, radiative scattering from surrounding mountains, or the ambiguity between surface and subsurface melting. 
\blue{We fixed the threshold to 3dB for inferring all targets, noting that \citet{johnson2020evaluation} found only minor differences between using 2dB and 3dB when evaluating various melt-detection methods against SAR-derived melt. Moreover, \citet{li2024glacier} also reported minor differences in Greenland melt days with adjustments of ±0.3dB to the 3dB threshold. However, \citet{li2024glacier} found a substantially higher melt frequency when using HV polarization instead of HH and attributed this difference to a deeper penetration depth or a higher sensitivity to low amounts of melt. Nevertheless, the HH estimates were generally in better agreement with weather-station-derived melt estimates. }
While improving the calibration of SAR-derived meltwater estimates remains an important research direction, our work addresses the complementary challenge of developing spatiotemporal interpolation techniques.

There are many machine learning techniques which may further improve the interpolation: For example, our UNet is not capable of accurately predicting events that occur on small temporal \textit{and} small spatial scales, such as the rapid drainage of small (\textless250m) lakes~\citep{miles2017}, as they are unresolved in the daily input data streams and the model is deterministic. Generative methods, such as diffusion models, have the potential to overcome this issue by learning distributions between a low-resolution input and multiple possible high-resolution targets~\citep{li22srdiff}.
Furthermore, the UNet inputs exclude possibly relevant information, such as atmospheric patterns or the timestamps of SAR observations, which could be overcome by more complex inputs and spatiotemporal model architectures. As evaluating the full suite of possibilities and novel ML methods goes beyond the scope of this study, we encourage MeltwaterBench to be used for intercomparing advances in meltwater downscaling algorithms.

The UNet has almost 50M parameters, which stands in contrast with the single parameter of the time interpolate SAR baseline. \blue{This difference in model complexity reflects a trade-off in interpretability, computational cost, and accuracy rather than a single best choice. For applications that require full interpretability or tolerate the smoothing of extreme melt events, as visualized in~\cref{fig:meltwater_extent_per_day}, the SAR-based interpolation, which retains considerable accuracy, may be the best choice. Where capturing fast transitions or overall accuracy matters most, the deep learning-created product performs best and can be used as an additional data stream in meltwater analysis.} \strikeolive{The lower parameter count greatly simplifies the interpretability. And, given the considerable accuracy of the SAR-based baseline it may be sufficient to use a purely SAR-based product for some applications. We remind that operational SAR data is not available at daily temporal scales, meaning that the running mean interpolation smooths out extreme melt events, as visualized in~\cref{fig:meltwater_extent_per_day}. However, in general we recommend to use the deep learning-created product as one data stream among several.}

\label{sec:temporal_generalization}
\label{sec:spatial_generalization}
Our proposed benchmark evaluates a model's ability to fill in spatiotemporal gaps in meltwater observations, but does not evaluate a model's capability to forecast, hindcast, or generalize to other locations. \blue{In particular, the specific model weights of our trained UNet are likely geographically limited because} \strikeolive{In particular,} meltwater within our study area primarily accumulates in the snowpack, which results in large areas that are detected as meltwater. In contrast, on the southwestern GrIS meltwater accumulates in ponds and streams, meaning a model trained on the southeastern GrIS would not necessarily generalize to these new conditions. Also, a generalization to Antarctica, such as the Larsen C ice shelves, would need to include the importance of winds.

\blue{While the trained weights may not transfer, many of the benchmark's insights likely generalize across regions, such as the time interpolate SAR being a strong and simple baseline (\cref{tab:results_overview}). Its methodology — including the preprocessing pipeline, baselines, and metrics — could also be repeated to re-establish such insights in new regions. Moreover, the benchmark serves as a testing ground that decouples model architecture development from data preprocessing, letting researchers get started on model development without first having to assemble the dataset.}

\subsection{Extensions of deep learning-based downscaling of meltwater}\label{sec:discuss_runoff}

Evaluating the deep learning method across climatic zones would also be necessary to extend our work from surface meltwater estimates to ice sheet mass loss. Estimating mass loss would further require scaling the computations to all of Greenland and establishing a link to annual runoff. Promisingly, the computational cost seems feasible based on~\cref{sec:hyperparameters} and a link to runoff may be achievable through statistical fits to point observations or RCM estimates of runoff volume~\citep{husman24meltwaterjames}.

We suggest two cases that may benefit from the high spatiotemporal resolution of our methodology: First, quantifying the drivers of regional rapid melt events. The UNet \blue{could be retrained as a forecasting model that depends on} winds and temperatures from multiple days preceding melt events. \blue{And, such a forecasting model could} facilitate sensitivity analyses between coastal melting and large-scale synoptic weather patterns, such as blocking events and atmospheric rivers. Second, the daily 100m information on meltwater across the Helheim glacier may assist with analyzing how local meltwater distribution impacts glacier flow and basal friction. Studies often rely on point measurements of meltwater~\citep{Stevens_Nettles_Davis_Creyts_Kingslake_Ahlstrom_Larsen_2022} and the gap-filled product may help contextualize those to melting in surrounding areas.

\section{Conclusion}\label{sec:conclusion}
 
We demonstrated that deep learning methods can improve on conventional algorithms for downscaling surface meltwater. In particular, deep learning can enhance regional climate model estimates of surface meltwater fraction by incorporating remote sensing datasets from SAR and PMW. Alternatively, a gridded product with notable accuracy (90\%) can also be created without deep learning through a running mean calculation with SAR data, but would underestimate extreme melt events. The deep learning-created daily 100m resolution maps of Helheim glacier's surface meltwater provide daily information in topographically complex areas and may aid in revealing insights about ice mass loss processes. More broadly, deep learning shows promise towards improving local biases in mass balance assessments and we hope MeltwaterBench can encourage further advances in downscaling algorithms.

%

%

\section{Reproducibility and data availability statement}
The code is published at \href{https://github.com/blutjens/meltwaterbench}{github.com/blutjens/meltwaterbench} \blue{with \href{https://doi.org/10.5281/zenodo.21890291}{doi:10.5281/ zenodo.21890291}} and associated with a CC BY-NC 4.0 license. The core dataset is published at \href{https://huggingface.co/datasets/blutjens/meltwaterbench}{huggingface.co/datasets/blutjens/meltwaterbench} \blue{with \href{https://doi.org/10.57967/hf/9969}{doi:10.57967/hf/9969}}, and the auxiliary dataset is distributed through the U.S. Antarctic Program Data Center (USAP-DC) at \href{https://www.usap-dc.org/view/dataset/601841}{usap-dc.org/view/dataset/601841}. Users of the core or auxiliary dataset are asked to reference this paper, as well as the data reference~\citep{HelheimData}. The core and auxiliary dataset are published with a CC BY-4.0 license. The created daily 100m product is available with CC BY-4.0 as \blue{doi-referenced} geotiffs on huggingface (2017-23), as a video (2018-23) at \href{https://youtu.be/OaonUT6dIbg}{youtu.be/OaonUT6dIbg}, \blue{and as an mp4-file on \href{https://doi.org/10.5281/zenodo.21905965}{doi:10.5281/zenodo.21905965}}.
Data from the Programme for Monitoring of the Greenland Ice Sheet (PROMICE) are provided by the Geological Survey of Denmark and Greenland (GEUS) at \href{https://promice.org/weather-stations/}{promice.org/weather-stations} under a CC-BY-4.0 license.

\authorcontributions
B.L. is the lead author and contributed to all CRediT author roles. P.A. processed the datasets and together with R.A. contributed to data curation, formal analysis, investigation, methodology, software, validation, and writing - original draft and writing - review and editing. T.W. contributed to software in the form of conducting the DeepLabv3+ experiment. G.C. contributed to conceptualization, formal analysis, funding acquisition, and writing - review and editing. M.T. contributed to conceptualization, funding acquisition, formal analysis, validation, and writing - review and editing.

\acknowledgments
This material is based upon work supported by the National Science Foundation Early-Concept Grants for Exploratory Research (NSF EAGER) under Grant No. (2136938). We appreciate USAP-DC, PROMICE, GEUS, ESA, Copernicus, NSIDC for creating openly accessible data products that significantly simplified our analyses. We are very grateful to Xavier Fettweis for providing the MAR data, and to Justin Kay and Matthew Kearney for conducting initial experiments on diffusion-based models. Thank you to Dava Newman for the help in funding acquisition and encouraging this work.

\appendix
\crefalias{section}{appsec}
\crefalias{subsection}{appsubsec}
\crefalias{subsubsection}{appsubsec}

\section{Additional information on data}\label{app:data}

\begin{figure}[t]
  \centering
      \includegraphics[trim=0 0 0 0, width=0.98\columnwidth, angle = 0]{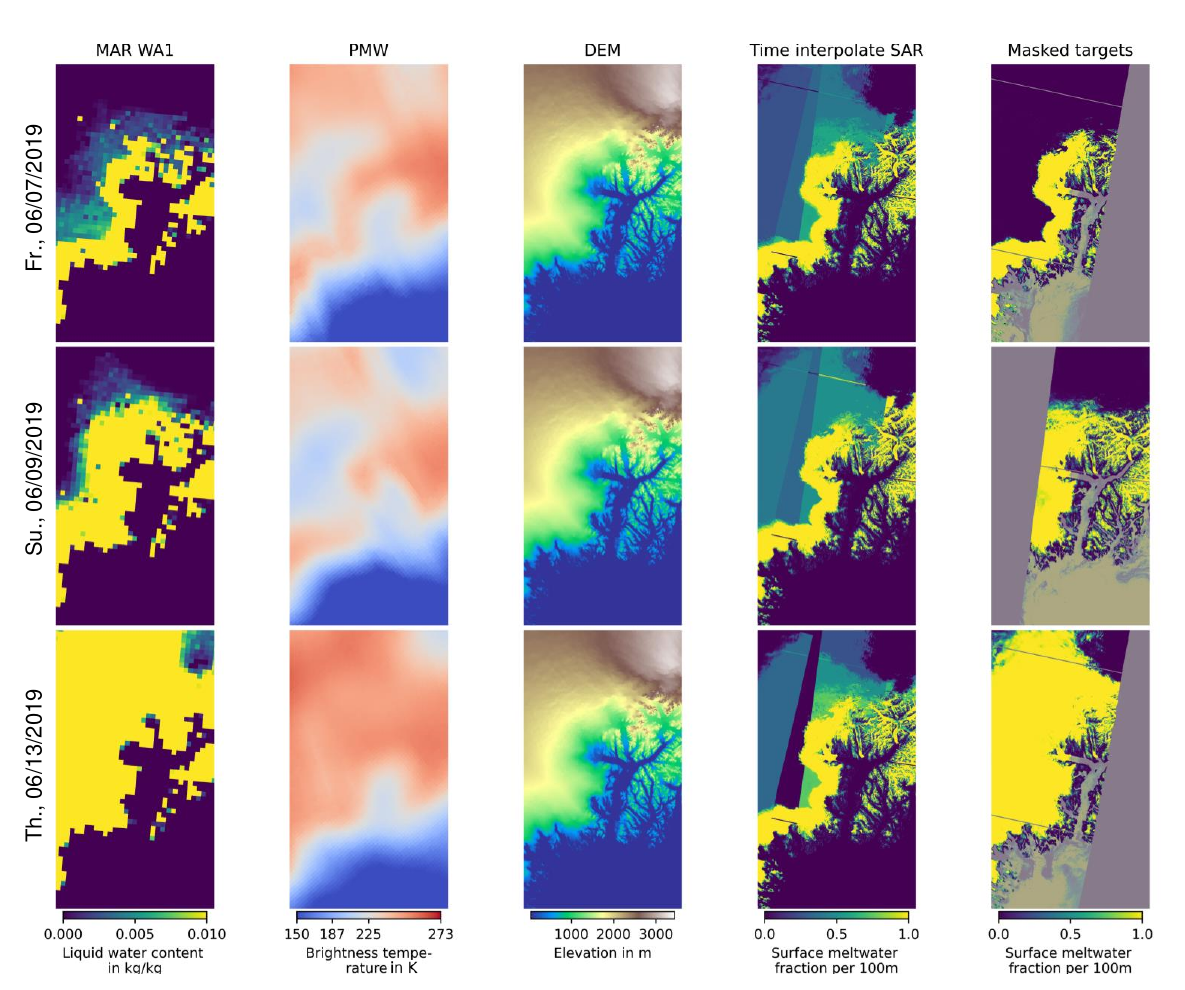}
\caption[]{Input data streams (left 4 columns) and target meltwater observations (right) during the 12th of June 2019 melt event. While the target observations have a revisit time of 1-12 days and contain significant masked areas, the input datastreams are available every day at every pixel.}
\label{fig:melt_event_2019_06_12} 
\end{figure}

\Cref{fig:melt_event_2019_06_12} shows the input data streams for the melt event on the 12th of June, 2019.

\subsection{Synthetic Aperture Radar (SAR) data}\label{app:sar}
\begin{figure}[t]
  \centering
      \includegraphics[trim=0 0 0 0, width=0.98\columnwidth, angle = 0]{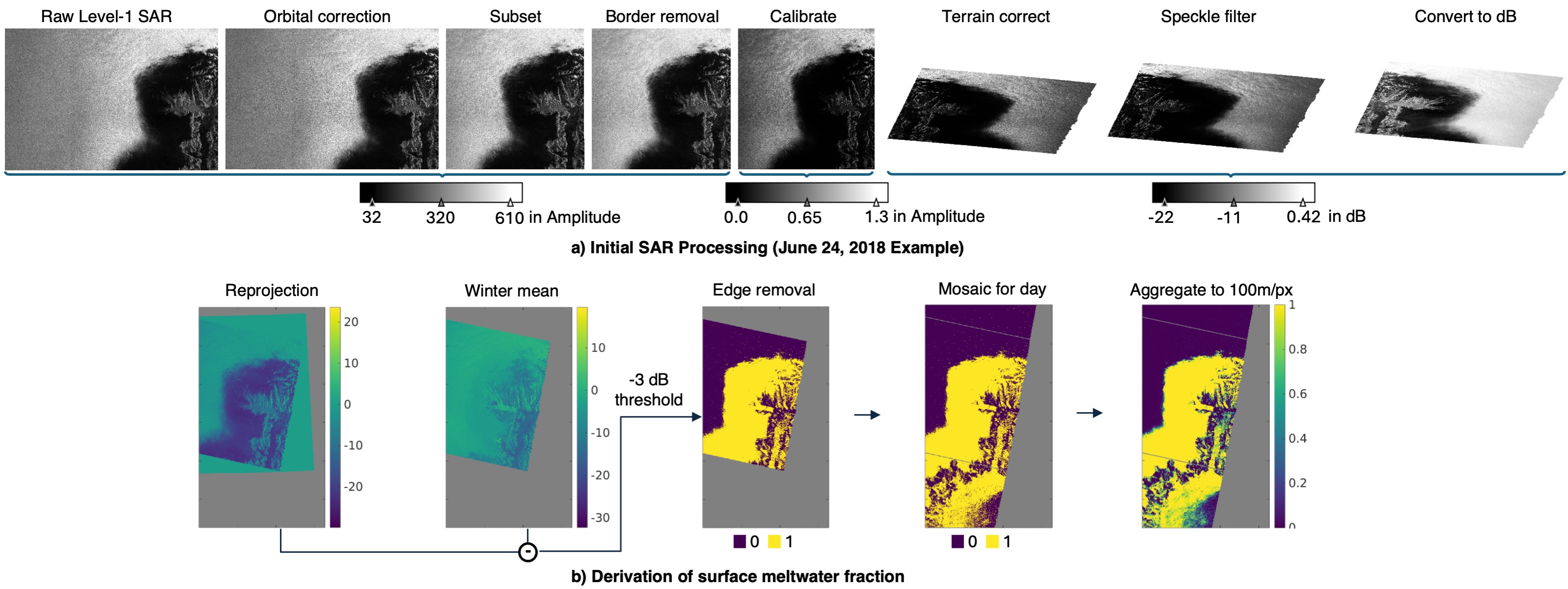}
\caption[]{\textbf{Deriving meltwater targets from SAR}. We preprocess SAR Level-1 data following the steps in (a). Then, we reproject and convert SAR backscattering intensity into surface meltwater fraction per 100m grid cell (b).}
\label{fig:sar_processing} 
\end{figure}

For reproducibility, we provide additional details on SAR processing here and depict our workflow in~\cref{fig:sar_processing}. The initial set of processing steps involved processing of raw level-1 data using the ESA Sentinel Application Platform (SNAP) graph processing (gpt) command line software~\citep{gptManual}. The SNAP processing involved utilizing built-in functions for orbital correction (with polynomial degree 3), subsetting within the domain of interest, border noise removal (with a border margin limit of 1500 pixels and threshold value of 0.5), radiometric calibration, speckle filtering (using a Lee Sigma filter with window size 7x7, sigma of 0.9, and target window size of 3x3), terrain correction, and conversion from linear to dB scale. These are standard procedures for processing SAR data \citep{filipponi2019sentinel} and the parameters chosen were default values with the exception of the terrain correction step and the border margin limit for border noise removal, which was increased to 1500 from 500 after testing on our input data. For the terrain correction step we utilized the 30m-resolved DEM, described in~\cref{sec:dem}. 
We reprojected the data onto a 10m equal area grid using bilinear interpolation.

During the final processing steps performed using MATLAB \citep{MATLAB}, we computed and subtract the previous year's per-pixel winter mean backscatter (Dec, Jan, Feb) from each 10m-resolved SAR scene and thresholded the data at -3dB such that any value below -3dB relative to the winter mean was considered to be melting. 
In order to ensure that variations in satellite orbit, viewing geometry, or data collection time did not introduce errors in the dataset, we only compared data for a given scene with winter data collected with the same orbital geometry. To ensure this, we grouped summer and winter data for a given year according to measurements collected at the satellite repeat time interval of 12 days. This was done by looping through files within the dataset and finding files within 12 days ± 4 seconds from the previous file. Following this procedure all winter files within a given group were averaged, and subtracted from the summer files within the same group to compute melting for a given summer image.

The initial processing steps were applied to $\sim$3000 observations. After computing melt, files were visually examined, revealing the persistence of some edge artifacts with apparently anomalous melting along the borders of the data swath in some images. This likely resulted from imperfections in the various corrections (i.e. terrain correction, orbital correction, and border noise removal) conducted for the original SAR data.  To remove this spurious data, we removed the borders for computed melt data using the following method:
\begin{enumerate}
    \item A binary mask was created to separate valid vs. missing data for the image.
    \item The canny edge detection method was applied to find the borders of valid data within the image using MATLAB software.  
    \item The edges were expanded using the ``diag'' and ``thicken'' functions.  
    \item Thickening was performed until a visual examination revealed that the edge artifacts had been removed.
\end{enumerate}

Following edge removal, all images from a given day were mosaicked using MATLAB to produce 10m resolution melt masks for individual days. In the case of overlap between images, if any of the images showed melt for a given pixel, that pixel was classified as melting. Melt data on the 10m grid were then aggregated onto the common 100m grid by computing the fraction of 10m grid cells exhibiting melt within each 100m grid cell. The 10m grid was chosen to be a subset of the 100m grid. The 100m fractional melt dataset was then used within the experiments for training and validation purposes. 

\subsection{Passive Microwave (PMW) data}\label{app:pmw}
We utilized enhanced resolution PMW data in the 37 GHz channel from the Special Sensor Microwave Imager/Sounder (SSMIS) sensor onboard the Defense Meteorological Satellite Program (DMSP) F17 satellite, available twice daily (with a morning and evening pass) at a spatial resolution of 3.125 km through the MEaSUREs program, distributed by the National Snow and Ice Data Center (NSIDC) \citep{Brodzik_data}.~\citet{husman24meltwater} uses the analogous 6.25km SSMIS product from the same source. We restrict our dataset to the evening pass ($\approx$18:30 local solar time~\citep{rs12142197}) for consistency.  The standard resolution for distributing PMW data is 25 km, obtained using a drop-in-the bucket method where multiple PMW measurements falling within a grid cell on a polar stereographic grid for a given day are averaged \citep{rs12162552}. The enhanced resolution product \citep{rs12162552} differs in that it provides a local measurement for morning and evening passes on an equal area grid, and enhances the spatial resolution by utilizing a different signal processing method in which information about the sensor measurement response function for measurements overlapping in space at a given overpass time is used to synthesize a higher resolution signal, enabling a higher level of detail in Greenland melt estimates \citep{colosio2020surface}. Despite the higher resolution grid, the effective resolution associated with the smallest resolvable feature is lower (between 3.125 and 25 km) given that a smoothing function is applied during processing \citep{rs12162552}.  

For our study, PMW data were obtained in the form of netcdf files from the NSIDC access tool. To produce the input dataset, raw netcdf files were converted to geotiff format using a shell script and the gdal-translate command line tool~\citep{rouault_2024_11175199}.  Finally, the geotiffs were reprojected to the Albers equal area 100m grid over the Helheim Glacier region using the gdalwarp command and nearest-neighbor interpolation.

This dataset along with other PMW data were used recently to produce a high-resolution record of Greenland ice sheet melt covering the period 1979-2019 using a threshold-based approach~\citep{colosio2020surface}.

If a PMW observation of a given day contains any missing values, we exclude this day from our train, val, and test datasets. A total of 35 images for which matching SAR observations are present were eliminated from the analysis. 
Our auxiliary dataset contains all data between Jan. 1, 2016 and Dec. 31, 2023 to be able to calculate the winter means that are needed for common PMW-based downscaling methods (see~\cref{sec:baselines}). 

\subsection{MAR regional climate model (RCM) data}\label{app:mar}

The MAR data is distributed in the form of yearly\blue{ densely gridded} netcdf files containing daily average model output. While we downloaded daily average output, data would also be available for download as temporal snapshots. A subset of variables, detailed in~\cref{tab:mar_data_o} were extracted from the raw data using netcdf Operators (NCO) software. The netcdf data were then converted to geotiff format and reprojected to the 100m equal area grid using gdal software and nearest-neighbor interpolation as was done for the PMW data. 

\begin{table}
    \caption{MAR data overview. Overview of all data variables in our auxiliary input dataset from MARv3.14. The core dataset contains daily averaged liquid water content within the top meter of snow (WA1) at 5km spatial resolution.}
    \label{tab:mar_data_o}
    \centering
    \resizebox{0.99\textwidth}{!}{
    \begin{tabular}{llllll}
    \toprule
         Variable&  abbr&unit&  Variable&  abbr& unit\\
         \midrule
 average liquid water content within the top meter of snow& WA1& kg/kg& surface albedo& AL2&-\\
         surface air temperature (2m above sfc.)&  TTZ&°C&  cloud optical depth&   COD& -\\
         specific humidity (2m above sfc.)&  QQZ&g/kg&  lower atmosphere cloud fraction&   CD& -\\
         y-direction (~northward) 2m wind speed&   V2Z&m/s&  middle atmosphere cloud fraction&   CM& -\\
 x-direction (~eastward)  wind speed& U2Z& m/s& upper atmosphere cloud fraction& CU&-\\
         rainfall&  RF&mm/day&  latent  heat flux&   LHF& W/$m^{2}$\\
 snowfall (in water equivalent units)& SF& mm/day& sensible heat flux& SHF&W/$m^{2}$\\
 melt & ME& mm/day& surface atmospheric pressure& SP&hPa\\
 surface mass balance (in water equivalent units)& SMB& mm/day& shortwave downward welling radiation& SWD&W/$m^{2}$\\
         &  &&  longwave downward welling radiation&   LWD& W/$m^{2}$\\
    \bottomrule
    \end{tabular}}
\end{table}

The auxiliary variables in~\cref{tab:mar_data_o} could add information, such as heatwaves, heavy rainfall, higher humidity, and sunny days (as indicated by high SWD) that can accelerate melting~\citep{hermann2020lagrangian,tedesco2020unprecedented,antwerpen2022assessing,ryan24griscloudradiativefeedback}. Low temperatures and high winds can accelerate refreezing. The auxiliary variables might also enable insight into larger spatial patterns~\citep{mioduszewski2016atmospheric,mattingly18grisatmosphericriver}, such as a high pressure area over southeasternmost Greenland and could indicate clockwise winds over Greenland that pull warm air from lower latitudes~\citep{lindsey22septheatwave}.

The MARv3.14 model domain encompasses the Greenland ice sheet as illustrated in~\cref{fig:studyarea}. MAR simulates surface and atmospheric processes within this regional domain while being forced with ERA5 data at the lateral and ocean surface boundaries. MAR has been used in several studies of Greenland surface processes~\citep{fettweis2017reconstructions,antwerpen2022assessing,delhasse2024coupling}.

\subsection{Digital elevation model (DEM) data}\label{app:dem}

As an additional input to the models discussed below, we used the Greenland Ice sheet Mapping Project (GrIMP) v2.0 digital elevation model (DEM) mosaic of the ice sheet surface topography, available at a 30m spatial resolution \citep{grimp2022data,howat2014greenland}. The product is derived from panchromatic stereoscopic imagery collected by the Maxar GeoEye-1, WorldView-1,-2, and -3 satellites, collected over the May 2008 through November 2020 period, and registered to ICESat-2 ATL06 lidar-derived elevations collected in the summers of 2019 and 2020 using the co-registration procedure of \citep{levinsen2013improving} \citep{grimpuserguide}. The product is distributed as a set of 36 tiles covering the Greenland ice sheet, in Geotiff format, and \blue{densely gridded over our study area}. To produce the DEM used as part of our analysis, we utilized the gdalwarp tool to mosaic the GrIMP tiles covering our study area, reproject them to the 10m resolution sub-grid of the common 100m resolution grid over the Helheim region, and then average the 10m data onto the common 100m resolution grid.

\subsection{Land-ocean mask}\label{app:grimp}

A similar procedure was conducted on the GIMP (earlier acronym for GrIMP) land-ocean mask \citep{gimp2017data,howat2014greenland}. The GIMP land-ocean mask was mapped using a combination of USGS Landsat 7 ETM+ panchromatic band and the RADARSAT-1 SAR images from the Canadian Space Agency. The mask is provided in GeoTIFF format similarly to the GrIMP DEM, at a 15, 30 or 90m spatial resolution. We chose the 30m data to be consistent with the DEM. As for the DEM, we used the gdalwarp tool to mosaic tiles overlapping with our study area, reprojected the data onto the 100m sub-grid, and averaged the mask onto the common 100m resolution grid.

\subsection{Additional information on data statistics and split}\label{app:data_statistics}\label{app:data_split}
We excluded the last month within our study period (Sep 2023) from the core dataset because MAR data was only available until Aug 31, 2023.

\begin{figure}[t]
  \centering
      \includegraphics[trim=0 0 0 0, width=0.98\columnwidth, angle = 0]{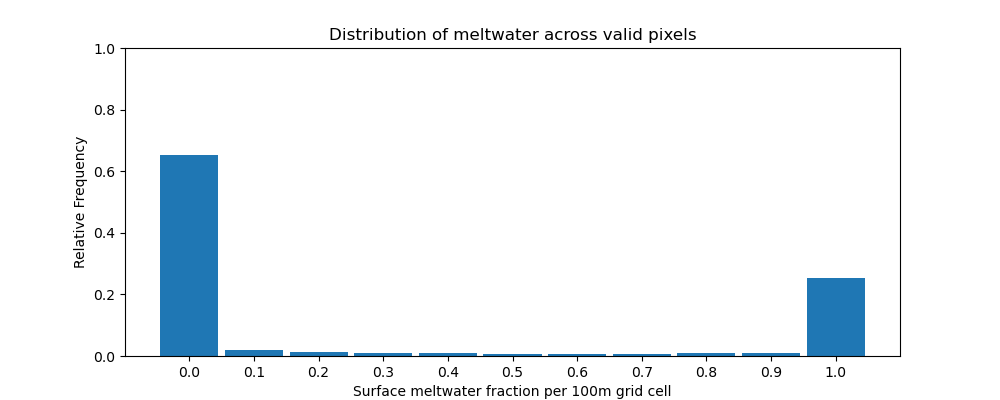}
\caption[]{Histogram of the surface meltwater targets across the training dataset. The targets are in the set of real numbers in $[0,1]$ and are displayed as a binned distribution ($n=11$) normalized to relative frequency. The targets are slightly imbalanced towards no-melt pixels (no-melt:melt = $65.4\%:34.6\%$), considering only valid pixels over land, and assuming a cut-off for binary classification at 0.10.}
\label{fig:histogram_meltwater}
\end{figure}

\Cref{fig:histogram_meltwater} shows the distribution of surface meltwater across the training dataset.

\begin{figure}[t]
  \begin{subfigure}{1.\textwidth}
      \centering
      \includegraphics[trim=0 0 0 0, width=0.98\columnwidth, angle = 0]{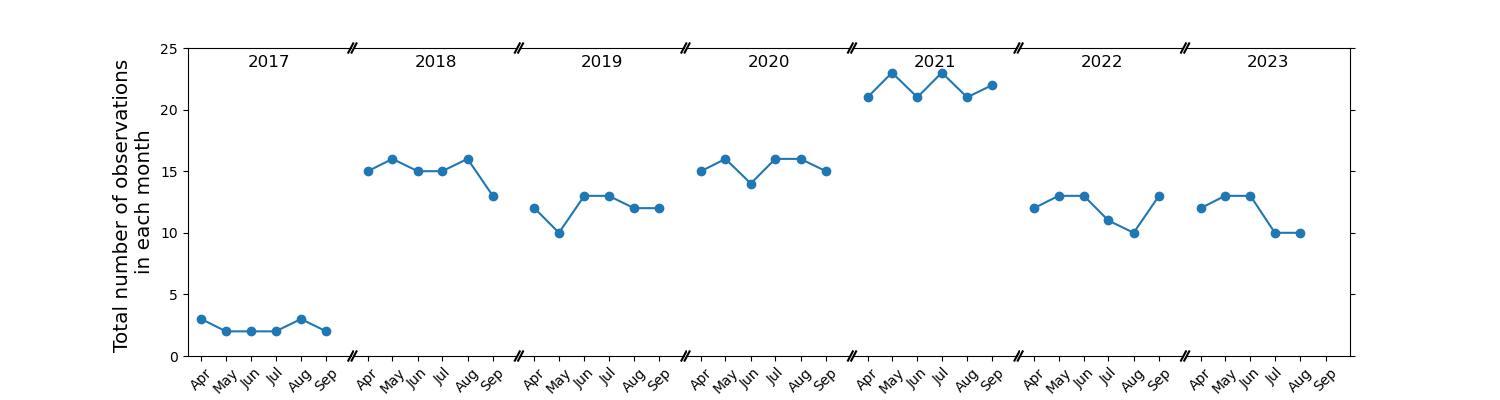}
      \caption[]{Total number of observed days per month. We count a day as observed if all core input and target data streams are available. Starting in 2022, data from S1B is unavailable.}
      \label{fig:total_observations_per_month} 
    \end{subfigure}
    \begin{subfigure}{1.\textwidth}
        \centering
        \includegraphics[trim=0 0 0 0, width=0.98\columnwidth, angle = 0]{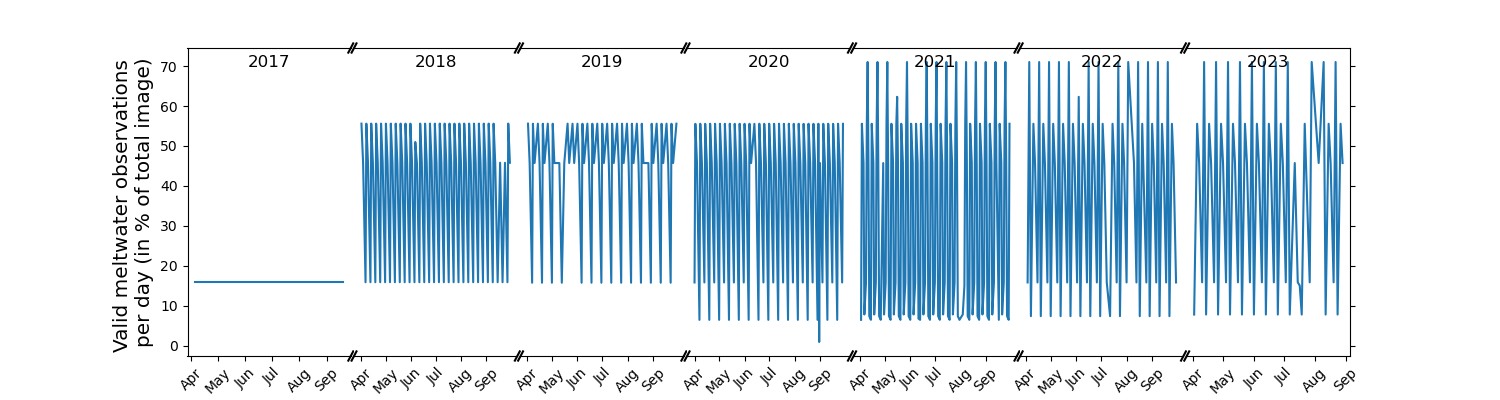}
        \caption[]{Valid target pixels per day as percentage of all pixels in the full-scale image. A target pixel is considered valid if the SAR-derived surface meltwater fraction at this pixel has been observed and has not been masked by the land-ocean mask or processing artifacts. On most days ${\approx} 5, 15, 45, 55,$ or $70\%$ of pixels are valid which corresponds to the S1A\&-B swath paths that are intersecting with our study area. The plot shows all images in the core dataset.}
        \label{fig:valid_observations_per_day}
    \end{subfigure}
    \caption{Additional statistics of the observed target meltwater fraction from SAR. During 2017, only a single satellite retrieval path overlaps with our study area (bottom), resulting in few observations (top).
    }
    \label{fig:target_statistics_per_day}
\end{figure}

\Cref{fig:total_observations_per_month} shows the total number of observed days per month and~\cref{fig:valid_observations_per_day} shows the observed pixels in \% of the total image per day across the entire dataset.

Our core dataset is arguably small to medium-sized in comparison to common super-resolution datasets: The total number of pixels (valid and invalid) in our dataset is equivalent to ${\approx}2,400$ images at  1K resolution, i.e., $(1024\times1024)$ pixels, which are less pixels than in the 1000 DIV2K images at $\approx$2K resolution, $30,000$ CelebA-HQ images at 1K resolution,
or 10-60 million LSUN images at 0.25K resolution~\citep{agustsson17div2k,karras18celebahq,yu16lsun}. In comparison to the ${\sim}20$GB of our core dataset, DIV2K-default measures ${\sim}4.7$GB, CelebA-HQ-1k ${\sim}54$GB, and LSUN-total ${\sim}1.1$TB~\citep{tf24datasets}. 

We reject the year 2017 from the validation and test sets.~\Cref{fig:total_observations_per_month} shows that the dataset contains 10-23 observations per month, except for 2017 which only records 2-4 monthly observations. Furthermore, all images in 2017 are only valid over the southwest of the glacier. Thus, we cannot robustly evaluate the accuracy of a model over the full study area for 2017. We keep the 2017 data in the training set to have additional training samples.

\textbf{R-squared.}
We calculate the R-squared, $R^2$, due to its common use. The R-squared is related to the MSE through a normalization factor proportional to a measure of dispersion and a rescaling to $[-\infty,1]$ -- with $1$ as the best value.

\section{Additional information on metrics and models}\label{app:metrics_and_methods}
\subsection{Appendix to Evaluation Metrics}\label{app:metrics}
The precision and and recall are calculated as follows.

\begin{equation}
\text{Prec}(\tY, \hat \tY) = \frac{1}{N_\text{valid}} \sum_k^K 
n_{\text{valid},k} 
\frac{\sum_{(i,j)\in IJ_{\text{valid},k}} \mathds{1}(y_{k,i,j} > y_\text{thresh}) \mathds{1}(\hat y_{k,i,j} > y_\text{thresh})}
{\sum_{(i,j)\in IJ_{\text{valid},k}} \mathds{1}(\hat y_{k,i,j} > y_\text{thresh})}
\label{eq:precision}
\end{equation}

\begin{equation}
\text{Rec}(\tY, \hat \tY) = \frac{1}{N_\text{valid}} \sum_k^K 
n_{\text{valid},k} 
\frac{\sum_{(i,j)\in IJ_{\text{valid},k}} \mathds{1}(y_{k,i,j} > y_\text{thresh}) \mathds{1}(\hat y_{k,i,j} > y_\text{thresh})}
{\sum_{(i,j)\in IJ_{\text{valid},k}} \mathds{1}(y_{k,i,j} > y_\text{thresh})}
\label{eq:recall}
\end{equation}

The F1-score is the harmonic mean of the computed precision and recall values:

\begin{equation}
\text{F1}(\tY, \hat \tY) = \frac{2}{\frac{1}{\text{Prec}(\tY, \hat \tY)} + \frac{1}{\text{Rec}(\tY, \hat \tY)}}
\label{eq:f1_score}
\end{equation}

\textbf{R-squared.} We calculate the $R^2$ score for a set of masked image tiles as follows:
\begin{equation}
R^2(\tY, \hat \tY) = \frac{1}{N_\text{valid}} \sum_k^K n_{\text{valid},k}
\max \left(-1, \left( 1 - 
\frac{\sum_{(i,j)\in IJ_{\text{valid},k}}{(y_{k,i,j} - \hat y_{k,i,j})^2}}
{\sum_{(i,j)\in IJ_{\text{valid},k}}{(y_{k,i,j} - \overline y_{k})^2}} \right)\right)
\label{eq:r2}
\end{equation}
where the max$(\cdot)$ term calculates the $R^2$ of tile $k$ and clips the $R^2$ of each tile $k$ to $[-1,1]$ to avoid large negative outliers that would skew the mean. The $n_{\text{valid},k}$ term reweights the value, such that, each valid pixel is given the same weight across all tiles; $\overline y_k$ is the mean over the target tile, 
$\overline y_k = \frac{1}{|IJ_{\text{valid},k}|}
\sum_{(i,j)\in IJ_{\text{valid},k}}{y_{k,i,j}}$.
If all pixels in the image tile are invalid, the $R^2$ of that tile is zero.

\subsection{Model set-up: Time interpolate SAR}\label{app:running_mean_sar}
The time interpolate SAR model forecasts every pixel to take the average values of the $k_h$ meltwater observations in the training set before and after at that location. If the prior or posterior image pixel is masked then the image value is not considered. This leads to a model that calculates the target value as:

\begin{equation}
\hat y_{k,i,j} = \frac{\sum_{k^\prime\in K_k} \mathds{1}\{(i,j)\in IJ_{\text{valid},k^\prime}\} y_{k^\prime,i,j}}
{\sum_{k^\prime\in K_k} 
\mathds{1}\{(i,j)\in IJ_{\text{valid},k^\prime}\} }
\label{eq:time_interpolate_sar}
\end{equation}
where $\hat y_{k,i,j}$ is the predicted pixel value for the $k$-th image at location $(i,j)$; $K_k = \{k-k_h,...,k-1,k+1,...,k+k_h\}$ is the set of image indices $k_h$ steps before and after the index $k$; $\mathds{1}\{(i,j)\in IJ_{\text{valid},k^\prime}\}$ is 1 if the location $(i,j)$ is in the set of valid pixels for the $k^\prime$-th image and 0 otherwise; $y_{k^\prime,i,j}$ is the pixel value of the $k^\prime$-th image at location $(i,j)$. \blue{We set any predicted pixel for which no valid observations exist across all $K_k$ images to zero. }

We choose the interpolation horizon, $k_h=3$. We treated this choice as a hyperparameter and found the best value on the validation set calculating scores for the set $k_h\in\{1,2,3\}$. With this horizon, the running mean equals a running mean over a span of $2-3$ weeks. $k_h=3$ was likely the best value because it ensures that there is at least one valid pixel at every location in most predicted images, but at the same time the surrounding images are not too far away from the target image. When using this model to predict an image, $\hat y$, in the validation dataset, the input images, $y$, are chosen only from the training dataset. We also use this model to generate the 'running mean' inputs for the machine learning models, in which case, both $\hat y$ and $y$ are drawn from the training dataset. 

\blue{This approach is inspired by previous studies which have used simple interpolation methods to fill gaps in SAR backscatter data, for instance, using the most recent SAR observation \citep{li2024glacier,luckman2014surface}, temporal interpolation \citep{wang2007melt}, and spatiotemporal interpolation \citep{ashcraft2006comparison}. Our approach is somewhat different in that here we fill gaps with meltwater fraction data rather than filling gaps in SAR backscatter data.}

\subsection{Model set-up: Interpolate MAR}\label{app:interpolate_mar}
The interpolate MAR model smoothes out the MAR WA1 inputs using a Gaussian Blur and then readjusts the cutoff value for melt/no-melt using brightness and gamma adjustment. Finally, a landmask is applied to mask out ocean pixels. We implement each transform using pytorch and find the best parameters for each by sweeping over the values in the following ranges: Gaussian blur kernel size $[91,201]$, Gaussian blur standard deviation $[33,99]$, gamma $[0.001,30]$, brightness factor $[40,200]$.

\blue{This method is inspired by previous studies that apply a threshold to MAR liquid water content in the top 1 m of the snowpack to produce a binary melt estimate in a manner consistent with observational data~\citep{fettweis2011melting,datta2018melting,dethinne2023sensitivity}. While we do not apply a threshold here, the method has a similar effect, with the added benefit of smoothing discontinuities between adjacent model pixels and producing a melt fraction rather than a binary estimate.}

\subsection{Model set-up: Threshold PMW}\label{app:threshold_pmw}
The threshold PMW model follows the implementation in~\citep{colosio2020surface}, eq. (3), which is based on the work by~\citep{tedesco09pmwthreshold}. The model computes the predictions, $\hat y_{k,i,j}$, as follows:

\begin{align}
    \hat y_{k,i,j} &= 
        \begin{cases}
          1, & \text{if}\ x_{k,i,j}^{\text{PMW}} > x_{\text{yr}_k,i,j}^{\text{PMW-Threshold}} \\
          0, & \text{otherwise}
        \end{cases}  \\
    x_{\text{yr}_k,i,j}^{\text{PMW-Threshold}} &= \gamma x_{\text{yr}_k,i,j}^{\text{PMW-winter}} + \omega\\
    x_{\text{yr}_k,i,j}^{\text{PMW-winter}} &= \frac{1}{\lvert\sK_{\text{yr}_k,\text{winter}}\rvert} 
        \sum_{k^\prime \in \sK_{\text{yr}_k,\text{winter}}} x_{k^\prime, i, j}^{\text{PMW}}
\end{align}

where $x_{\text{yr}_k,i,j}^{\text{PMW-Threshold}}$ is the brightness temperature threshold (in Kelvin) at location $(i,j)$ for the year $\text{yr}_k$ of the $k$-th image. 
The threshold is a linear function of the Winter mean brightness temperature, $x_{\text{yr}_k,i,j}^{\text{PMW-winter}}$, with intercept, $\omega$, and slope $\gamma$ which are constants based on an electromagnetic model as detailed in~\citep{colosio2020surface}. 
The Winter mean is the average per-pixel value over all images in January and February of the corresponding year; denoted by the indices $k^\prime \in \sK_{\text{yr}_k,\text{winter}}$.

Tuning the parameters $\gamma$ and $\omega$ on the training dataset might result in a more accurate model for our use-case. However, we decided against tuning this model to maintain comparability with prior comparative studies, e.g., in~\citep{colosio2020surface}. We use the values $\gamma=0.48$ and $\omega=128K$ that are reported in
~\citep{colosio2020surface} which finds the best parameters across Greenland. The biases do not necessarily represent what the bias across Greenland looks like and is rather to point out possible discrepancies when zooming into one location in detail.

\subsection{Model set-up: Threshold DEM}\label{app:threshold_dem}
The threshold DEM model finds the best monthly linear fit and cut-off values to the DEM input. The model predictions are

\begin{align}
\hat y_{k,i,j} &= \text{tanh}_\text{hat}(a_{\text{mon}_k} x^{\text{DEM}}_{i,j} + b_{\text{mon}_k}) \; \text{with} \\
\text{tanh}_\text{hat}(z) &= 0.5 \left(\text{tanh}(c+z) + \text{tanh}(c-z) \right)
\label{eq:linear_dem}
\end{align}

where $\text{mon}_k\in \{4,...,9\}$ indices the month of the $k$-th image, $x_{i,j}^\text{DEM}$ is the DEM input at location $(i,j)$, and $a_{\text{mon}_k}, b_{\text{mon}_k}$, are parameters and $c$ is a hyperparameter. The $\text{tanh}_\text{hat}$ is a custom activation function that resembles a boater straw hat -- it is zero except for a continuous one-valued region with tanh-shaped transitions. The parameters $a_{\text{mon}_k}$ and $b_{\text{mon}_k}$ are independent of the year and determine the width and location of the threshold region while $c$ determines the sharpness of zero-one transitions. The hyperparameter $c$ is set to $c=4$ and the best parameters $(a_{\text{mon}_k},b_{\text{mon}_k})$ are found using the training and validation dataset after 22 epochs with stochastic gradient descent, batch size $16$, MSE loss, initial learning rate $10$, and reducing it by $\times 0.1$ every $5$ plateauing epochs. The total number of free parameters is $12$ due to  $a_{\text{mon}_k}$ and $b_{\text{mon}_k}$ each taking a different scalar value for $6$ months (Apr.-Sept.).

The lower threshold value should correspond with the firn line, below which the ice and snow completely melt during the Summer leaving a surface of bare rock
; also called ablation zone.
\blue{
\subsection{Model set-up: random forest}\label{app:random_forest}
The random forest uses a single pixel from PMW, MAR, time interpolate SAR, and the digital elevation model as input features. Appending two time features in the form of a sin and cos of the normalized day of year did not improve evaluation scores, so we exclude time features. We train the random forest over a randomly sampled subset of 20\% of all pixels in the training set and evaluate it on 20\% randomly without replacement drawn  pixels from the validation set. Using less data did not significantly impact evaluation scores (up to 1\% of data) and using more data caused memory issues. To find the best hyperparameters, we first fix the number of estimators to 200, the minimum number of samples of each leaf to 4, the minimum number of samples to split to 4, and the number of CPU cores to 22. Then, we sweep over the ratio of bootstrapped samples in each estimator [0.01,0.05,0.1,0.25,0.5], the number of features [2,4], and the maximum depth of each tree [4,8,16,20,None]. We optimize the random forest using MSE loss due to the significantly higher computational cost of optimizing random forests on $L_1$ loss. Then, we select the best hyperparameter combination as a trade-off between MSE and $L_1$ loss on the validation set and training time. The best hyperparameters were a bootstrap ratio of 0.05, the number of features 2, and a maximum depth of 16. The best random forest has 5944362$\approx$5.94M split thresholds, i.e., 30K per estimator, which we report as the total number of parameters.
}

\subsection{Model set-up: vanilla UNet}\label{app:unet_vanilla}

For all deep learning model experiments we normalize each input channel to zero-mean, unit-variance. During early experiments the predictions contained sharp edges which came from the nearest-neighbor interpolation of the large-scale PMW and MAR inputs, so we smooth these inputs using a Gaussian blur filter with kernel size 45 and 99 and standard deviation of 15 and 33 pixels, respectively. We do not normalize the targets and predictions. Data is kept at float32 precision. We use a sigmoid activation in the last layer to bound the predictions to the physically plausible range, $[0,1]$.

\begin{figure}[ht]
  \centering
      \includegraphics[trim=0 0 0 0, width=0.98\columnwidth, angle = 0]{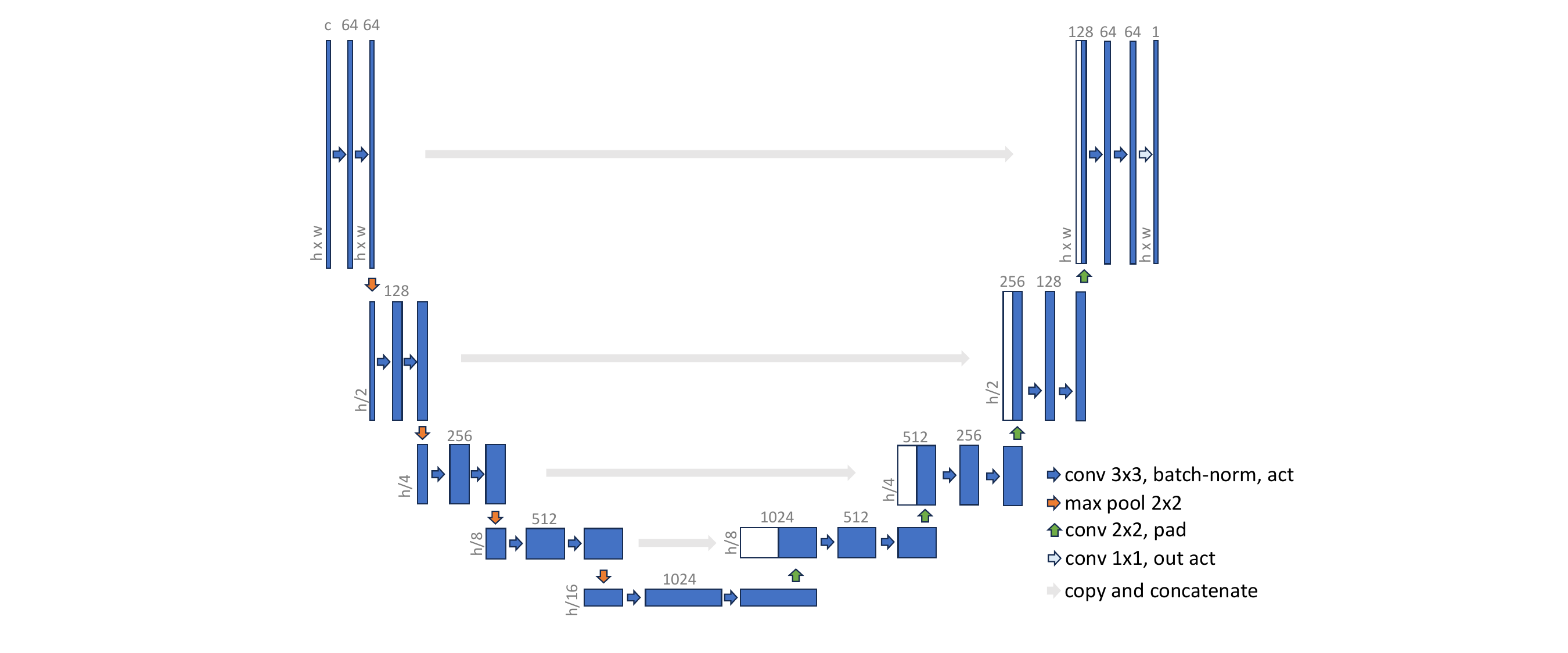}
\caption[]{\textbf{Vanilla UNet architecture.} The vanilla UNet combines an encoder-decoder architecture with skip connections. The numbers beside and above the blue embedding layers represent the spatial and feature dimension, respectively. The feature dimension is constant in between layers for which the depicted layer width remains constant.}
\label{fig:unet_architecture}
\end{figure}

Our vanilla UNet has $31,038,209\approx31.0$M weights occupying 120MB at float32 precision.~\Cref{fig:unet_architecture} illustrates the model architecture, number of layers, blocks, and features. The reported results were trained for 500 epochs, batch size 8, tile size $h=w=512$, adam optimizer, weight decay $0.00001$, learning rate $0.0001$, GELU activation, sigmoid out activation, $L_1$ loss, and a learning rate scheduler reducing by $\times 10$ on plateaus with a patience of $100$. The weights are trained from scratch and initialized with the pytorch default Kaiming uniform initialization with negative slope $\sqrt{5}$~\citep{kaiming15initialization}.

Choosing the tile size is a trade-off: we presume that a large size of input tiles enables the vanilla UNet to identify and correct large spatial biases in, e.g., the MAR inputs, increasing the accuracy. The size of the theoretical receptive field (TRF) of the chosen vanilla UNet architecture is $(188\times188)$px (see calculation below) and depends on the number of encoding blocks (here 4), kernel size, and size of the max pool. Thus, the vanilla UNet can theoretically correct biases of the MAR inputs within a radius 18.8km (=188px*100m/px), which intersects a maximum of $(5\times5)$ MAR pixels of $5km/px$ resolution.   
In general, the quality of CNN predictions is known to degrade towards the image borders due to the increasing influence from border padding. Given the TRF, any output pixels within $188$px of the image border could theoretically contain edge artifacts. Thus, increasing the tile size also increases the number of pixels that are not affected by edge artifacts. The downside however is that larger tile size increases computational cost and can increase the required data for training~\citep{wang18pix2pixhd}. After initial experiments (unshown), we determined a tile size of $512$ as a good trade-off. 

The size of the theoretical receptive field (TRF) can be computed layerwise by starting with the center-pixel of the output and counting how many pixels in the previous layers it connects to. For the decoder, the output center-pixel connects to $4$ pixels in the final $32\times32\times1024$ feature vector. For the encoder, every $3\times3$ convolution adds $2$ pixels to the receptive field and every $2\times2$ max pool doubles the receptive field. For example, the center-pixel connects to $4+2+2=8$ pixels in the $32\times32\times512$ feature vector; connects to $8*2+2+2=20$ pixels in the $64\times64\times256$ feature vector; connects to $20*2+2+2=44$ in the $128\times128\times128$ feature vector ; ...; and connects to $92*2+2+2=188$ pixels in the input layer which is the TRF. This calculation could also be written as $4*2^b+\sum_{b^\prime=0}^b 4*2^{b^\prime}$ with $b=\{0,1,...4\}$ reverse-indexing the blocks. 
The effective receptive field is smaller than the TRF, because the number of connections per input pixel to the output center-pixel decreases with the distance from the input center-pixel~\citep{loos24receptivefield}.

\subsection{Model set-up: UNet SMP}\label{app:unet_smp}
Our optimized UNet SMP has $49,569,081\approx49.6$M weights occupying 189MB at float32 precision. The model encoder is fully detailed in Sec. 3.2 and Fig. 4~\citep{chen18deeplabv3plus} as `modified aligned xception', X-71. This encoder's main component are 22x xception blocks that each contain 3x depthwise-separable convolution blocks and a skip connection. Each of the depthwise-separable convolution blocks applies a depthwise convolution, batch normalization, ReLU, pointwise convolution, batch normalization, and ReLU. Separating the convolution across the horizontal (depth) and channel (point) dimension allows for deeper networks while maintaining a similar number of parameters. The encoder reduces the spatial dimension by a factor of $32\times$ by using convolutions with stride $=2$. \blue{Owing to the additional blocks and the $32\times$ (instead of 16x) spatial reduction, the UNet SMP has a larger receptive field than the $188$px of the vanilla UNet. }We do not use atrous spatial pyramid pooling in the UNet SMP encoder. The encoder is pretrained on ImageNet-1k which is a classification task on RGB data. We initially expected that these weights would not transfer well to our multi-modal dataset~\citep{zhu24geospatialfoundation}, but saw clear gains in the validation SSIM score in ablation experiments to randomly initialized weights. The decoder is similar to~\citep{ronneberger15unet}, uses five resolution levels, and is trained from scratch.

The reported results were trained with batch size $8$, tile size $h=w=512$, adam optimizer, no weight decay, $18$ workers, batch norm in the encoder but not in the decoder, ReLU activation, and a sigmoid activation in the output layer. Our scores improved by using a `CosineAnnealingWarmRestarts' scheduler that reduces the learning rate from an initial, $0.0001$, to $0.$ within the first cycle of $T_0=35$ epochs, and resets the learning rate afterwards. We train for a total of $1085$ epochs restarting this cycle $5\times$ and increasing $T$ by $2\times$ after every restart~\citep{loshchilov17cosineannealingwarmrestarts}. 
Using GELU marginally improved scores in the vanilla UNet, but is unavailable in the reference codebase so we use ReLU.

\subsection{Model set-up: DeepLabv3+}\label{app:deeplabv3plus}
Our DeepLabv3+ model is the model described in~\citep{chen18deeplabv3plus}, including atrous spatial pyramid pooling, and we use the implementation in segmentation models pytorch~\citep{qubvel23segmodels}. The DeepLabv3+ model is also a CNN-based encoder-decoder architecture, but in comparison to the vanilla UNet, has $30\%$ more parameters, only one skip connection, and a larger receptive field by using atrous convolutions with different rates in the bottleneck layer~\citep{chen18deeplabv3plus}. The model has 42,9M weights occupying 164MB at float32 precision. 

We swept over the encoder size (tu-xception41p vs. tu-xception71), encoder output stride (8,16), learning rate (0.001,0.0001,0.00001), decoder atrous rates ([6,12,18],[12,24,36], [24,48,72]), and weight decay (0.0001,0.). 
Both model sizes, tu-xception41p with 28.0M and tu-xception71 with 42.9M parameters, showed similar performance while we found tu-xception71 to have slightly better SSIM and tu-xception41p with slightly better MAE, MSE, and R2. We selected the hyperparameters that achieved the highest SSIM score on the validation dataset. 
The reported model uses the parameters: epochs $1085$ (stopped at $1081$), batch size $16$, tile size $h=w=304$, encoder tu-xception71, initial learning rate $0.001$, weight decay $0.$, adam optimizer, ReLU activation, sigmoid out activation, encoder depth $5$, encoder output stride $8$, decoder channels $256$, decoder atrous rates $[6,12,18]$, $4x$ upsampling, we use batch normalization, a cosine annealing scheduler with warm restarts and $T_0=35$, $T_\text{mult}=2$, $\eta_\text{min}=0.$, and masked $L_1$-loss. No data augmentations were applied. The network weights are randomly initialized. To generate large-scale tifs with this model, we use a prediction stride of $s_\text{val}=s_\text{test}=240$, and an erode size of $e_\text{val}=e_\text{test}=32$ which, paired with the $304$px tile size, results in no overlap between each tile. 

During the hyperparameter sweep we found that a higher learning rate, larger model, and no weight decay increased the performance. We observed no conclusive trends in varying the encoder output stride or decoder atrous rates.
DeepLabv3+ was trained for 1065 epochs taking 21 hours. In preliminary experiments, we also trained the SR3 diffusion model~\citep{saharia23sr3} which required significantly longer training times.

\section{Additional Results}\label{app:results}

\subsection{Additional metrics}\label{app:extra_metrics}
\Cref{tab:extra_metrics} shows additional metrics for each traditional method, UNet, and DeepLabv3+. The RMSE is computed as:
\begin{equation}
    \text{RMSE}(\tY,\hat\tY)=\sqrt{\text{MSE}(\tY,\hat\tY)}
    \label{eq:rmse}
\end{equation}
with the same unit as the targets (surface meltwater fraction per 100m grid cell). The PSNR is computed as:
\begin{equation}
    \text{PSNR}(\tY,\hat\tY) = 10 \text{log}_{10}(s_{\text{max}}^2 / \text{MSE}(\tY,\hat\tY))
    \label{eq:psnr}
\end{equation}
with the maximum image value $s_\text{max}=1$, and using decibels as the unit. 

We compute the standard deviation across images for the spatial error metrics, $\text{Err}_s$, weighted by the amount of valid pixels in each image. This provides a measure of the error variance across images rather than across valid pixels.
\begin{equation}
    \sigma(\tY,\hat \tY) = \sqrt{\frac{1}{N_\text{valid}} \sum_k^K 
    n_{\text{valid},k} 
    \left[
    \frac{1}{n_{\text{valid},k}} \sum_{(i,j)\in IJ_{\text{valid},k}} (\lvert y_{k,i,j} - \hat y_{k,i,j}\rvert)^p - 
    \text{Err}_s(\tY,\hat \tY)
    \right]^2}
    \label{eq:metric_std}
\end{equation}

The test-val score difference is computed by, first, computing the score for each model on the test set and on the val set. Then, we take the difference between the test and val set score for each model. Lastly, we compute the absolute value of this difference and, then, take an average across all models.

\begin{table}[t]
  \caption{\textbf{Results table.} Mean and standard deviation of additional evaluation metrics for each model across the test dataset. The standard deviation is computed according to~\cref{eq:metric_std}, best scores are \textbf{bold}, and we report three significant digits. UNet SMP is abbreviated as UNet.
}
  \centering
  \resizebox{0.99\textwidth}{!}{
  {
  \begin{tabular}{llllllllll}
    \toprule
    Model     &PSNR$_s$&RMSE$_s$ &    $R^2$  
&Prec.&Rec.&$\sigma_{\text{MAE}_s}$&$\sigma_{\text{MSE}_s}$  &$\sigma_{\text{Acc.}}$&$\sigma_{\text{SSIM}}$\\
    \midrule
    Time interpolate SAR &14.1&0.197 &    0.579
&0.755&0.880&0.0425
&0.0346
  &0.0679&0.0878
\\
    Interpolate MAR &8.276&0.386 &    -0.0518
&0.753&0.593&0.0902
&0.084
  &0.0992&0.179
\\
 Threshold PMW&5.959& 0.504 &    -0.464
&0.662&0.480&0.149
&0.145
  &0.148&0.189
\\
    Threshold DEM&8.638&0.370 &    0.103
&0.750&0.654&0.106
&0.0926
  &0.103&0.165
\\
    \blue{Random Forest} &15.2&0.174 &    0.652
&0.739&\textbf{0.906}&0.0352
&0.0244
  &0.0422&0.0865
\\
    DeepLabv3+      &14.8&0.182 &    0.648
&0.846&0.815&0.0285&0.0207  &0.0287&0.0812\\
    UNet&\textbf{16.0}&\textbf{0.158} &    \textbf{0.735}
&\textbf{0.866}&0.831&0.0252
&0.0170
  &0.0267&0.0747
\\
    Vanilla UNet&14.5&0.189 &   0.603
&0.814&0.868&0.0322
&0.0271
  &0.0319&0.0784
\\
\hline
    test-val score diff.      &0.268&0.00775 &    0.0164
&0.0117&0.0282&n/a&n/a  &n/a&n/a\\
    \bottomrule
  \end{tabular}}
  }
  \label{tab:extra_metrics}
\end{table}

\subsection{Meltwater extent per day}

\begin{figure}[H]
  \centering
    \subfloat[Vanilla UNet and traditional methods]{
      \includegraphics[trim=3cm 0 3cm 0, width=0.97\linewidth, angle = 0]{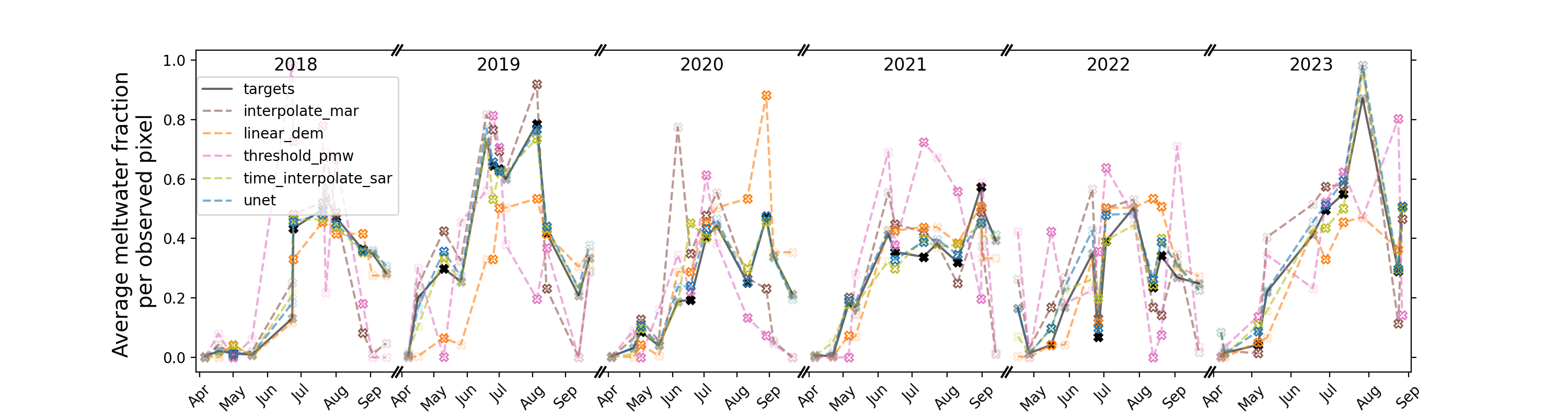}
\label{fig:unet_and_trad_avg_meltwater_preds_per_day_test}
    }  
    \\
    \subfloat[UNet SMP, DeepLabv3+, \blue{and random forest}]{
      \includegraphics[trim=3cm 0 3cm 0, width=0.97\linewidth, angle = 0]{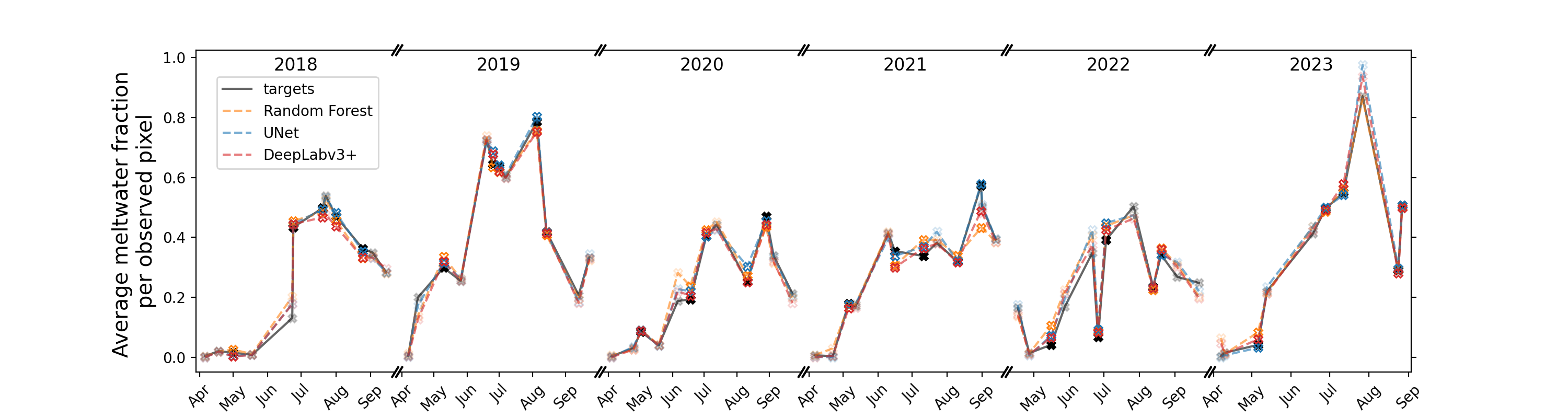}
\label{fig:unet_and_deeplabv3_avg_meltwater_preds_per_day_test}
    }  
\caption[]{\textbf{Predicted and observed daily surface meltwater fraction on the test set.} The black line shows the observed surface meltwater fraction as an average across all pixels that were valid and observed on the given day. The same valid-pixel mask is applied to the predictions before plotting the surface meltwater fraction average. The X-marker is shaded on days with $<20\%$ valid pixels and solid otherwise. Threshold DEM is plotted as linear\_dem.}
\label{fig:avg_meltwater_preds_per_day_test} 
\end{figure}

\begin{figure}[t]
  \centering
    \subfloat[UNet SMP and only SAR]{
      \includegraphics[trim=3cm 0 3cm 0, width=0.97\linewidth, angle = 0]{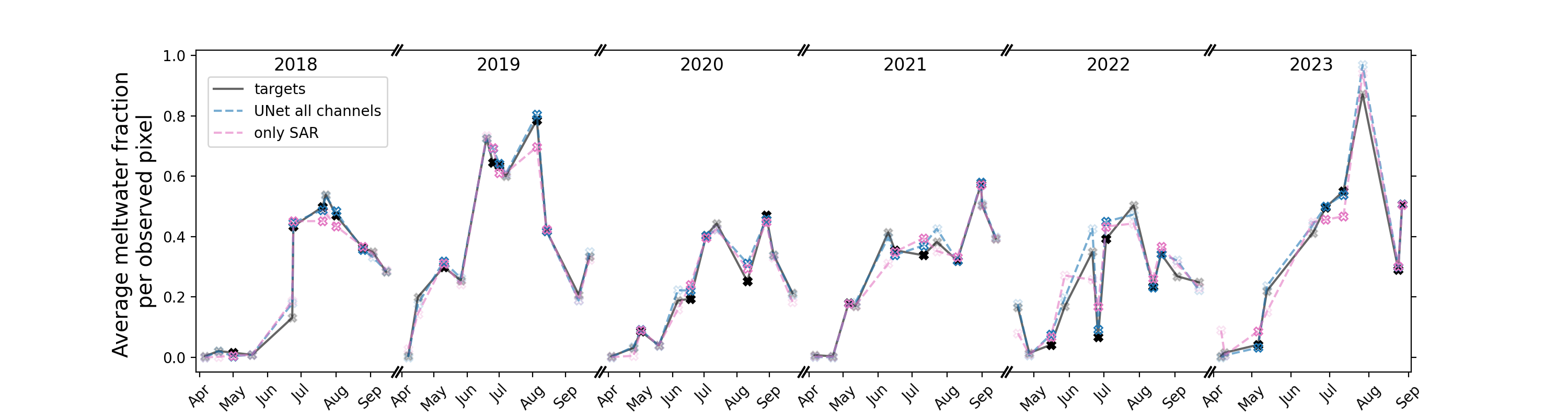}
\label{fig:unet_and_only_sar_avg_meltwater_preds_per_day_test}
    }  
\caption[]{\blue{\textbf{Daily surface meltwater fraction for observations and a UNet using only SAR inputs.} The black line shows the observed surface meltwater fraction as an average across all pixels that were valid and observed on each day in the test set (similar to~\cref{fig:avg_meltwater_preds_per_day_test}). The same valid-pixel mask is applied to the predictions from a the UNet SMP with all channels (blue) and a retrained UNet SMP that exclusively uses the SAR channel (pink). We point out the lower predictions of only-SAR with respect to UNet SMP on 2023-06-28 and 2023-07-12 which seem to be days of rapidly increasing meltwater (see~\cref{fig:meltwater_extent_per_day_unet}. The X-marker is shaded on days with $<20\%$ valid pixels and solid otherwise.}}
\label{fig:avg_meltwater_preds_per_day_test_only_sar} 
\end{figure}

~\Cref{fig:avg_meltwater_preds_per_day_test} and \cref{fig:avg_meltwater_preds_per_day_test_only_sar} were created by training each model on the training set and then comparing each model's predictions against every available observation in the test dataset. Because of the partial masks in each SAR observation we plot the average meltwater fraction per observed pixel. Some SAR observations contain valid pixels only in a small subsection of our study area meaning large deviations of the ML models are less representative of the full study area. Thus, we mark days for which $<20\%$ of pixels are valid by shading the X-marker.

\begin{figure*}[t]
  \centering
  \centering
  \begin{subfigure}{1.\textwidth}
      \centering
      \includegraphics [trim=0.5in 0 1.1in 0, clip, width=0.99\textwidth, angle = 0]{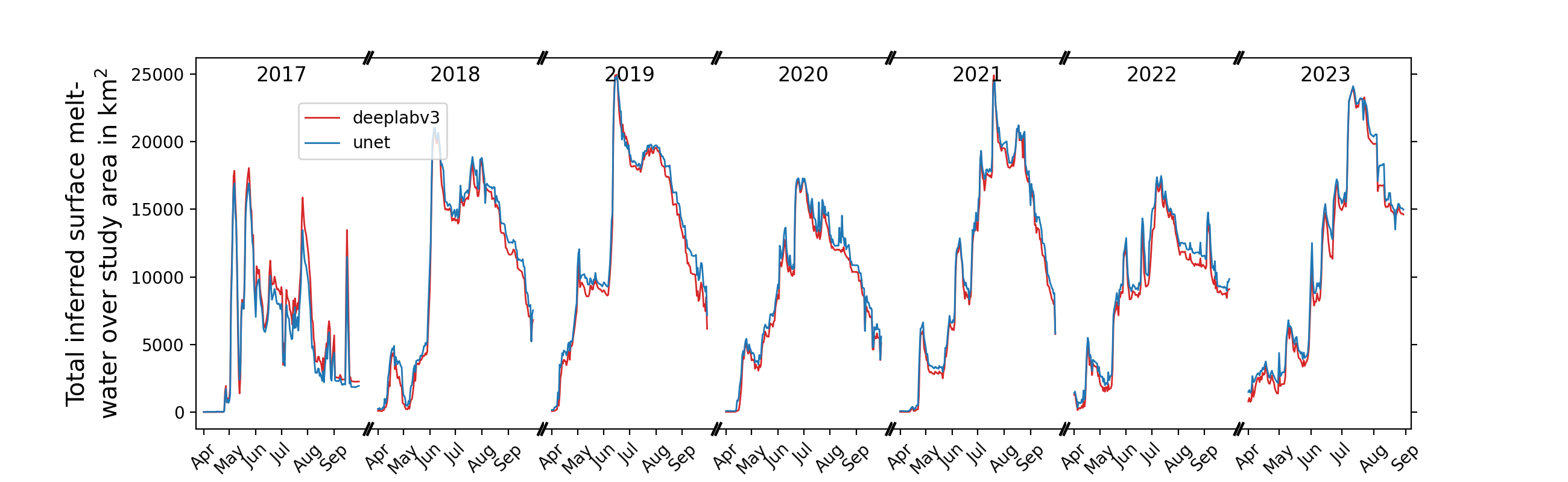}
  \end{subfigure}
\caption[]{Predictions of total cumulative surface meltwater per day by the vanilla UNet (blue) and DeepLabv3+ (red) for every day in the study period 2017-2023.
} 
\label{fig:meltwater_extent_per_day_unet_deeplabv3} 
\end{figure*}

~\Cref{fig:meltwater_extent_per_day_unet_deeplabv3} shows the daily predictions of total cumulative surface meltwater for the UNet SMP, DeepLabv3+, \blue{and random forest}. Notably, DeepLabv3+ predicts less meltwater in comparison to the vanilla UNet which is also reflected by the better precision and worse recall in~\cref{tab:extra_metrics}.

\subsection{Station observation analysis}\label{app:station_observation_analysis}

\begin{figure}[t]
  \centering
  \includegraphics[width=0.99\linewidth, angle = 0]{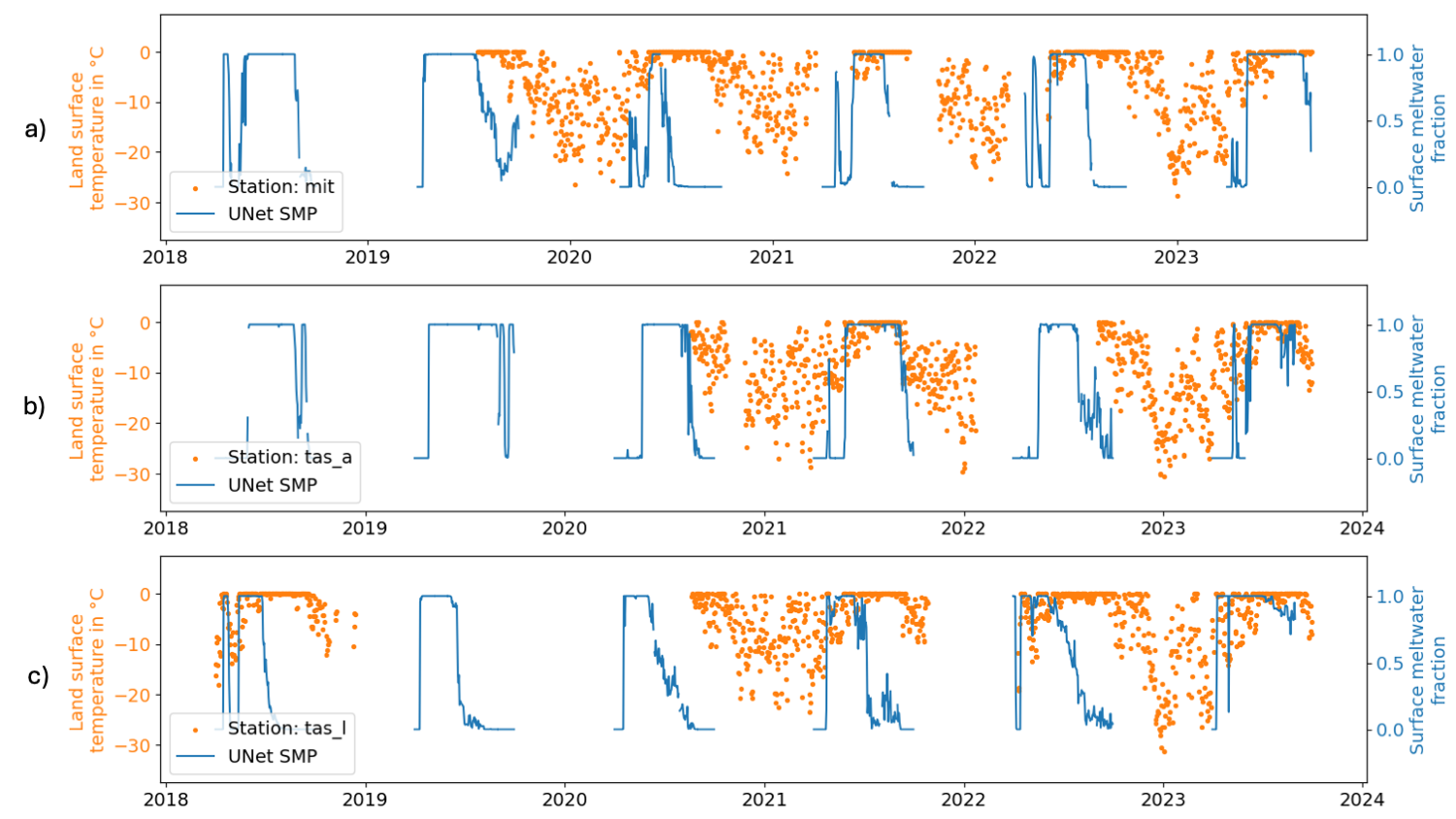}
\caption[]{\blue{\textbf{All station observations.} The figure shows every valid observation of land surface temperature for each station over 2018 to 2023 (orange, a: mit, b: tas-a, c: tas-l). The figure also shows the corresponding UNet SMP predictions of surface meltwater fraction over each station location (blue). The x-axis ticks indicate 1st Jan.}}
\label{fig:meltwaterbench_station_analysis_complete}
\end{figure}

\blue{~\Cref{fig:meltwaterbench_station_analysis_complete} shows every valid day's 6am land surface temperature (LST) observations of the PROMICE weather stations and corresponding UNet predictions from 2018 to 2023. The UNet predictions align well with the tas-a station LST observations in~\cref{fig:meltwaterbench_station_analysis_complete}b) 2021 and 2023, i.e., LST$\geq -1^\circ$C when UNet$==1$. However, the LST disagrees with the UNet during the late-seasons of the mit station in~\cref{fig:meltwaterbench_station_analysis_complete}a) 2020 and 2022, as well as the tas-l station in~\cref{fig:meltwaterbench_station_analysis_complete}c) 2018 and 2021. The station tas-a is more inland than tas-l and mit, indicating that the late-season disagreement on tas-l and mit may be due to coastal biases or exposed late-season rock. The weather stations contain many missing or NaN values which are not plotted and the UNet predictions have only been created over April 1st to Sept. 30th of each year.}

\subsection{Length of observational gap}\label{app:mae_over_observation_gap}

\begin{figure}
\centering
\includegraphics[width=0.99\linewidth, angle = 0]{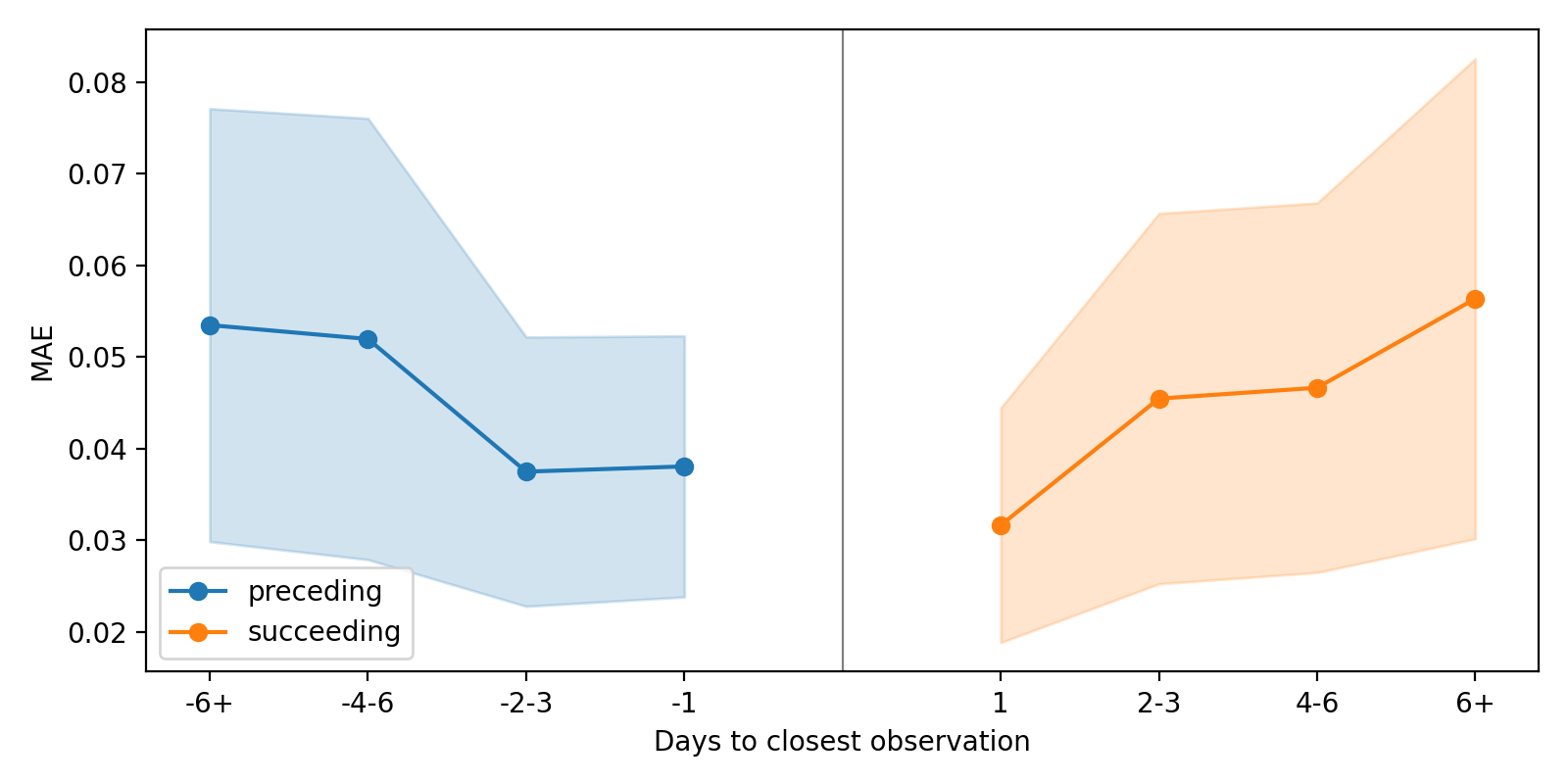}
\caption[]{\blue{\textbf{MAE over length of observational gap.} The figure shows the test set mean absolute error (MAE) of UNet SMP surface meltwater predictions as a function of the number of days to the closest preceding (blue) or succeeding (orange) observation. The absolute error and length of the observational gap are computed per-pixel, and the shaded area shows the variance of this absolute error. The closest `observation' to each pixel are the temporally closest valid SAR-derived targets in the training set that are also used in time interpolate SAR.}}
\label{fig:mae_over_observation_gap}
\end{figure}

\blue{In our spatiotemporal gap-filling task models are tasked to interpolate over short (1-3 day) and long (4-12 day) gaps in observations. To understand the decay of model accuracy over long observational gaps, we plot the mean absolute error (MAE) of UNet predictions over the length of the observational gaps:~\Cref{fig:mae_over_observation_gap} shows that the MAE worsens with increasing distance to the last (blue) or next (orange) observation, as expected. Moreover, the MAE only grows moderately, with the 6+ day observational gaps still achieving acceptable MAE. We binned longer observational gaps into 6+ as there are only few longer gaps and the error variance becomes large.
}

\subsection{Deep learning hyperparameter insights}\label{app:hyperparameters}

Overall the hyperparameter tuning improved the scores from $\sim0.715$ to $\sim0.765$ SSIM, but took significant effort.
We varied over model architectures [UNet, DeepLabv3+], backbone families [Xception, ResNet, ConvNeXt], learning rates [1e-6 to 1e-2], batch norm, pretrained vs. randomly initialized weights, a scheduler that reduces the learning rate when the loss curve plateaus vs. a cosine annealing scheduler with warm restarts~\citep{loshchilov17cosineannealingwarmrestarts}, tile size [32,64,128,256,512], activation function [ReLU, GELU], and the loss function [$L_1$, MSE]. Using batch norm stabilized our training, i.e., without batch norm some models with high learning rate ($>0.001$) diverged to poor SSIM values ($<0.4$) and activating batch norm remediated that. Using the cosine annealing scheduler slightly improved our results ($\sim 0.01$ SSIM on a UNet with Xception71 backbone) in an ablation with the plateau scheduler. The DeepLabv3+ predictions are slightly blurry, partially due to a final 4x upsampling layer that is not followed by any additional layers.
Reducing the prediction stride and increasing erode size during inference achieved only negligible improvement in SSIM ($\sim0.005$), but took significant extra compute.

\section{Extended Discussion}\label{app:discussion}

The biases of the threshold DEM model likely reflect the shift in distribution between train and test dataset, visualized in~\cref{fig:avg_meltwater_preds_per_month_test}b. In particular, we note that there is less meltwater during August in the test dataset in comparison to the training dataset and the threshold DEM model, only using static information as inputs, does not contain any information on this distribution shift.

\subsection{MeltwaterBench as testbed dataset for ML algorithm development}\label{app:ml_testbed}
We developed MeltwaterBench to evaluate data-driven downscaling algorithms in the context of surface meltwater. Nevertheless, the assembled benchmark might also be suitable for studying fundamental ML advances in the following areas:

\textbf{Physics-constrained ML.} Physics-informed ML downscaling methods are an active area of interest~\citep{lutjens24floodviz,abraham25physicsinformed}. A common assumption in physics-informed ML downscaling is that patch-based averages of the predicted high-resolution field equal the corresponding low-resolution pixel values~\citep{harder23physdownscaling}. However, this assumption is not valid between the MeltwaterBench inputs and targets, indicating that the benchmark poses novel challenges for embedding physical constraints into downscaling algorithms. An interesting physics-informed approach could be to consider surface mass balance or to split downscaling into a bias-correction and superresolution task.

\textbf{Foundation models.} Current benchmarks for geospatial foundation models do not include a downscaling task~\citep{lacoste23geobench, marsocci24pangaea}. However, geospatial foundation models could be highly applicable to the downscaling problem due to the need for data efficient models. In our case, for example, the size of the target dataset is limited by the number of retrievals during the S1A\&-B SAR satellite lifetime. And, even expanding the training dataset with meltwater observations over other locations could come with a considerable domain shift, for example, due to the increasing ponding of meltwater in Western Greenland or the wind-driven freezing over Antarctica's Larsen C Ice Shelf. Thus, geospatial foundation models that are pretrained on vast datasets to promise data-efficient fine-tuning could help~\citep{klemmer25satclip,jakubik25terramind}, and it would be an interesting outcome if they can improve on our task-specialized UNet baseline.

\bibliography{references}


\end{document}